\documentclass[10pt,journal,compsoc]{IEEEtran}

\usepackage{nicefrac}       
\usepackage{epsfig}
\usepackage{graphicx}
\usepackage{amsmath}
\usepackage{amssymb}
\usepackage{url}
\usepackage{amsthm}
\usepackage{algorithm}               
\usepackage{algorithmic}             
\usepackage{multirow}                
\usepackage{bm}
\usepackage{footmisc}
\usepackage{color}
\usepackage{booktabs}       
\usepackage{color}
\usepackage{xfrac}
\usepackage{subfigure}
\usepackage[T1]{fontenc}
\usepackage{pifont}
\usepackage{hyperref}       
\usepackage{flushend} 

\ifCLASSOPTIONcompsoc
  \usepackage[nocompress]{cite}
\else
  \usepackage{cite}
\fi
\ifCLASSINFOpdf
\else
\fi
\allowdisplaybreaks

\def\R{\mathbb{R}}

\def\l{\left}
\def\r{\right}
\def\inp{\emph{\emph{in}}}
\def\outp{\emph{\emph{out}}}
\def\outpp{\emph{\emph{out}}}

\def\SR{r^{\emph{sr}}}
\def\INR{\operatorname{\Phi_{{INR}}^{{Loc}}}}
\def\INRG{\operatorname{\Phi_{{INR}}^{{Glob}}}}
\def\ASSR{\operatorname{ASISR_{EQ}}}
\def\CNN{\operatorname{\Phi^{EQ}_{CNN}}}
\def\Encoder{\operatorname{\Phi_{Enc}^{EQ}}}
\def\MLP{\operatorname{MLP}}
\def\Bic{\operatorname{\phi_{Bic}}}

\def\outC{\check{\circ}_\psi}
\def\intC{\tilde{\circ}}
\def\inpC{\hat{\circ}_\varphi}
\def\({\l(}
\def\){\r)}
\def\[{\l[}
\def\]{\r]}
\def\red{}

\title{An Equivariant Proximal Operator for Deep Unfolding Methods in Image Restoration}

\author{%
  David S.~Hippocampus\thanks{Use footnote for providing further information
    about author (webpage, alternative address)---\emph{not} for acknowledging
    funding agencies.} \\
  Department of Computer Science\\
  Cranberry-Lemon University\\
  Pittsburgh, PA 15213 \\
  \texttt{hippo@cs.cranberry-lemon.edu} \\
}
\begin{document}

\title{Rotation Equivariant Arbitrary-scale Image Super-Resolution}
\author{ Qi Xie, Jiahong Fu, Zongben Xu and Deyu Meng
\thanks{This work was supported in part by the National Key R$\&$D Program of China under Grant 2024YFA1012000; in part by the China NSFC projects under Grant 62272375 and 62206214; in part by Tianyuan Fund for Mathematics of the National Natural Science Foundation of China under Grant 12426105. (Corresponding author: Deyu Meng.)}
\thanks{Qi Xie, Jiahong Fu and  Zongben Xu  are with the School of Mathematics and Statistics, Ministry of Education Key Lab of Intelligent Networks and Network Security, Xi'an Jiaotong University, Shaanxi, P.R.China (e-mail: xie.qi@mail.xjtu.edu.cn, jiahongfu@stu.xjtu.edu.cn, zbxu@mail.xjtu.eud.cn).
}
\thanks{Deyu Meng is with the School of Mathematics and Statistics and Ministry of  Education Key Lab of  Intelligent Networks and Network Security, Xi’an Jiaotong University,  Xi’an, Shaanxi, China, and Macao Institute of Systems Engineering, Macau University of Science and Technology, Taipa, Macao (e-mail: dymeng@mail.xjtu.edu.cn).}
}

\markboth{IEEE Transactions on Pattern Analysis and Machine Intelligence, 2025}
{Fu \MakeLowercase{\textit{et al.}}: An Equivariant Proximal Operator for Deep Unfolding Methods in Image Restoration}

\IEEEtitleabstractindextext{
\begin{abstract}
The arbitrary-scale image super-resolution (ASISR), a recent popular topic in computer vision, aims to achieve arbitrary-scale high-resolution recoveries from a low-resolution input image. This task is realized by representing the image as a continuous implicit function through two fundamental modules, a deep-network-based encoder and an implicit neural representation (INR) module. Despite achieving notable progress, a crucial challenge of such a highly ill-posed setting is that many common geometric patterns, such as repetitive textures, edges, or shapes, are seriously warped and deformed in the low-resolution images, naturally leading to unexpected artifacts appearing in their high-resolution recoveries. Embedding rotation equivariance into the ASISR network is thus necessary, as it has been widely demonstrated that this enhancement enables the recovery to faithfully maintain the original orientations and structural integrity of geometric patterns underlying the input image. Motivated by this, we make efforts to construct a rotation equivariant ASISR method in this study. Specifically, we elaborately redesign the basic architectures of INR and encoder modules, incorporating intrinsic rotation equivariance capabilities beyond those of conventional ASISR networks. Through such amelioration, the ASISR network can, for the first time, be implemented with end-to-end rotational equivariance maintained from input to output. We also provide a solid theoretical analysis to evaluate its intrinsic equivariance error, demonstrating its inherent nature of embedding such an equivariance structure. The superiority of the proposed method is substantiated by experiments conducted on both simulated and real datasets. We also validate that the proposed framework can be readily integrated into current ASISR methods in a plug \& play manner to further enhance their performance. Our code is available at  \href{https://github.com/XieQi2015/Equivariant-ASISR}{https://github.com/XieQi2015/Equivariant-ASISR}.
\end{abstract}

\begin{IEEEkeywords}
Arbitrary-scale super-resolution, rotation equivariant, implicit neural representation, equivariance error analysis.
\end{IEEEkeywords}}
\maketitle
\IEEEdisplaynontitleabstractindextext
\IEEEpeerreviewmaketitle

\newtheorem{Thm}{Theorem}
\newtheorem{Rem}{Remark}
\newtheorem{Lem}{Lemma}
\newtheorem{Cor}{Corollary}

\IEEEraisesectionheading{\section{Introduction}\label{sec:introduction}}

\IEEEPARstart{I}mage super-resolution (SR) seeks to enhance the resolution of a low-resolution (LR) image to obtain a high-resolution (HR) image, which has been a long-standing low-level computer vision task \cite{park2003super,elad1999super}.
In recent years, the advent of deep learning has provided a powerful tool for the SR task, and various DL-based SR frameworks have been developed \cite{dong2015image, yang2019deep, wang2020deep}. Among them, the arbitrary-scale image super-resolution (ASISR) task, aiming to achieve arbitrary-scale HR recoveries from an LR input image, is recently becoming one of the most interesting topics in the low-level computer vision community \cite{chen2021learning,lee2022local,song2023ope, cao2023ciaosr, xu2021ultrasr}.

\begin{figure}
    \centering
    \includegraphics[width=\linewidth]{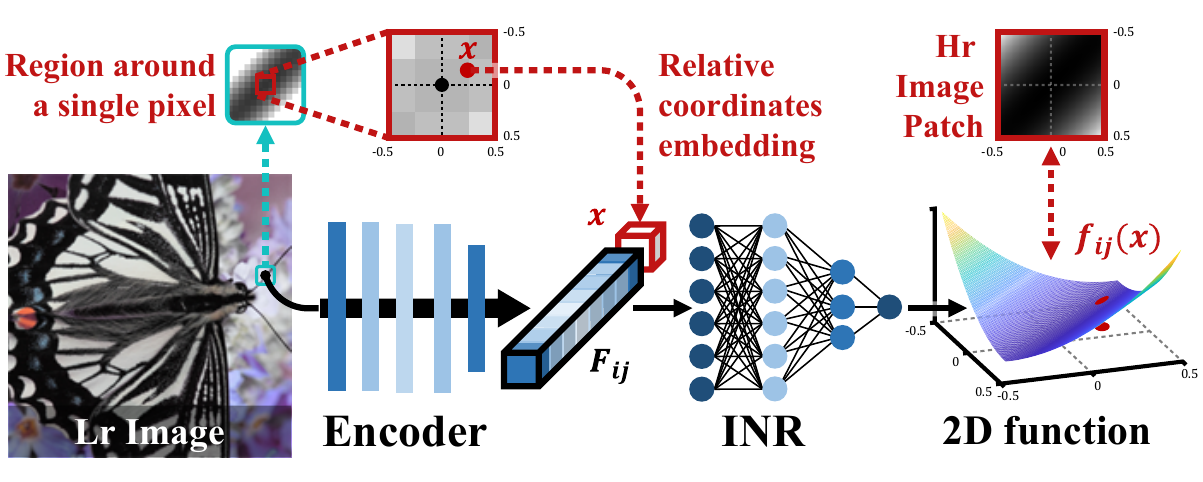}
    \caption{Illustration of the overall ASISR framework, where a continuous implicit function is predicted for each pixel of the input LR image.}
    \label{fig:fig1}\vspace{-2mm}
\end{figure}

Current ASISR networks are generally composed of two major components, including a deep-network-based encoder and an implicit neural representation (INR) module \cite{hu2019meta, chen2021learning, lee2022local,song2023ope}.
Specifically, as shown in Fig. \ref{fig:fig1}, the encoder is functioned to represent the region around each pixel of an input image as a set of latent codes, and then achieves a continuous implicit image function at an arbitrary position by inputting this coordinate, together with the corresponding local latent codes, into the subsequent INR module. Then, an arbitrary-scale super-resolution image can be naturally obtained, with its pixel values at the correlated coordinates calculated by this implicit image function.

Both components of ASISR have been carefully designed in previous research and play crucial roles in affecting its final recovery performance. For the design of deep-network-based encoder, CNN-based architecture has been typically adopted \cite{szegedy2015going}, attributed to its intrinsic translational equivariant property\footnote{Equivariance of a transformation in a network implies that applying the transformation to the input induces a predictable and consistent transformation of the intermediate feature maps and output, while preserving certain inherent properties of the system.}, which should be the most representative prior structure possessed by general images. Very recently, more advanced transformer-based architectures have also been attempted for this encoder design task, which further achieves performance enhancement, benefiting from its reasonable use of self-attention modules \cite{liang2021swinir}.

\begin{figure*}
    \centering
    \includegraphics[width=\linewidth]{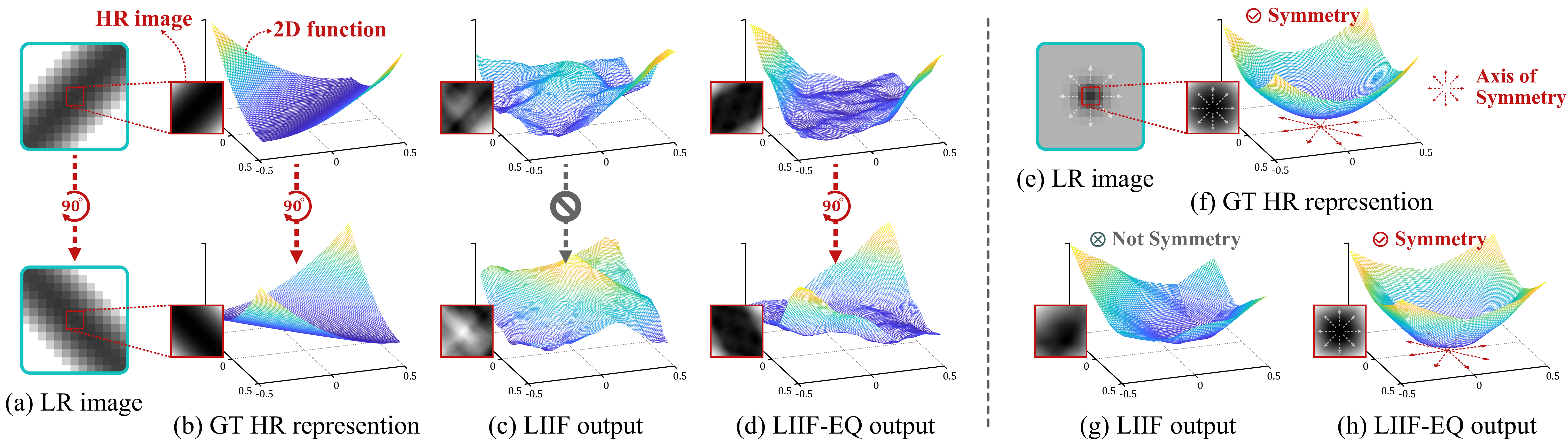}
    \caption{Illustration of the output local implicit image function obtained by LIIF and LIIF enhanced with the proposed method (LIIF-EQ). (a) two input LR images where the lower one is the $90^{\circ}$ rotated image of the upper one; (b) the correlated ground truth implicit image functions of two LR images; (c) and (d) the local implicit image functions achieved by LIIF and LIIF-EQ, respectively; (e) an input LR image with a local symmetry pattern; (f) the correlated ground truth implicit image functions; (g) and (h) the local implicit image functions achieved by LIIF and LIIF-EQ, respectively.}
    \label{fig:fun}
\end{figure*}

As a more distinctive element in ASISR framework, the INR module allows local image patches to be represented as 2D continuous functions, as shown in Fig. \ref{fig:fun}.
Its architecture was first designed as a multilayer perceptron (MLP) using the Local Implicit Image Function (LIIF) method \cite{chen2021learning}, by mapping coordinates and latent codes together to pixel values to achieve continuous image restoration for an LR input image. After that, more complex INR modules have also been proposed to further improve the performance or reduce parameters for ASISR, such as \cite{lee2022local,song2023ope,cao2023ciaosr,wu2021scale}.

\red{Despite achiving notable progress, ASISR remains in a significantly more ill-posed setting compared to conventional SR tasks. Under the requirement of recovering an arbitrary-scale high-resolution image with high quality, the LR images are with extreme information deficiency  and the current ASISR networks still struggle to effectively capture complex fine-grained structural representation within the LR images. This issue inevitably results in the emergence of numerous unexpected artifacts during reconstruction, often manifesting as severely warped and deformed geometric patterns, such as textures, edges, or shapes. }

\red{Therefore, compared with conventional SR methods, the  ASISR methods require more effective embedding of essential image priors into their network architectures. It should be noted that most existing ASISR methods are  constructed on black-box networks and heuristic image processing modules, little consideration has been given to the aforementioned crucial issue.}

Incorporating rotation equivariance \cite{Cohen2016group, weiler2019general, shen2020pdo,fu2024rotation} into the ASISR architecture is thus necessary, as it has been widely substantiated that this enhancement enables the recovery to more faithfully preserve the original orientations and structural integrity of the geometric patterns underlying the input image, even in severely degraded conditions \cite{lim2017enhanced, zhang2018residual,zhang2018image}. \red{Two typical instances on this aspect are intuitively shown in Fig. \ref{fig:fun}. 
One is that the embedding of rotation equivariance helps to better preserve nonlocal orientational similarity in SR reconstruction, as clearly shown by comparatively observing Fig. \ref{fig:fun}(c) and (d). This is because rotation equivariance leverages similarities along different orientations across non-local regions of the image, which ensures a more faithful recovery of image features like curves (with local similarity along smoothly varied directions) and sharp angles (with local similarity along directions with rapid variations) in HR output. Another is that rotation equivariance benefits a better preservation of local isotropic symmetry in HR recovery, as comparatively shown in Fig. \ref{fig:fun}(g) and (h). The above property ensures that features with similar characteristics in all directions, such as repetitive textures and isotropic patterns, can be consistently reconstructed without introducing directional artifacts. }

\red{Additionally,  theoretical researches have rigorously proven that equivariant models can attain  stronger generalization capabilities  for image processing tasks \cite{elesedy2022group,petrache2023approximation,sokolic2017generalization}, which is anticipated  to effectively alleviate the ill-pose issue of ASISR.}

However, designing an ASISR architecture that intrinsically integrates the rotation equivariance is actually a challenging task. Specifically, to address such a highly ill-posed task, a comprehensive redesign of the ASISR architecture is essential to effectively integrate and fully harness rotation equivariance within the network. That is, this redesign should ensure seamless preservation of end-to-end rotational equivariance, maintaining consistency throughout the network flow. \red{This consequently leads to difficulties in rotation-equivariant redesign for the INR module\footnote{The encoder part of ASISR can be relatively easier to be transformed into a rotation equivariant encoder by replacing traditional convolutions with rotation equivariant convolutions \cite{xie2022fourier}.},  since the INR modules in current methods are often constructed based on the coupling of deep network modules and complex operators that lack established rotation-equivariant counterparts \cite{chen2021learning,lee2022local,song2023ope,cao2023ciaosr,wu2021scale}.}

To address such a critical issue, this study proposes a novel manner for designing rotation equivariant (Rot-E) INR module for ASISR, as well as making advancements to ameliorate the design of the Rot-E encoder. The contribution of this work can be mainly summarized as follows:

(1) We first time propose the formulation for designing a Rot-E ASISR architecture with approximate end-to-end rotational equivariance maintained throughout the entire network flow. Specifically, we elaborately design Rot-E network modules for INR, including Rot-E input, intermediate, and output modules, to guarantee a thorough rotation equivariance embedding on this part. Besides, for the design of encoder in ASISR, we ameliorate a Rot-E convolution method based on the Bicubic representation, which further enhance the representation accuracy of current Rot-E convolutions (as shown in Fig. \ref{fig:bicubic} in Sec. 3.3). 

(2) We can readily reform previous ASISR architectures with additional rotation equivariance representation capability by employing the proposed Rot-E techniques. In particular, we have thoroughly validated that by ameliorating typical INRs for ASISR, including LIIF \cite{chen2021learning} (representing MLP-base INR), LTE \cite{lee2022local} (representing complex nonlinear function-based INR) and OPE \cite{song2023ope} (representing Non-parametric INR), into Rot-E versions through a plug-and-play (PnP) manner, their image representation ability and recovery performance can be readily improved consistently (which can be evidently observed in Fig. \ref{fig:fun}).

(3) We provide a comprehensive theoretical analysis to evaluate the intrinsic equivariance error for the proposed entire ASISR network, especially its INR part, demonstrating its inherent nature of embedding such expected rotation equivariance structure.

(4) We validate the superiority of the proposed method in experiments conducted on both simulated and real data in terms of rotation symmetry preservation capability, SR recovery accuracy, and generalization ability from natural images to hyperspectral images (HSI). Especially, attributed to its more sufficient parameter sharing essence, under the guarantee of fine recovery performance, the proposed method inclines to be with less network capacity. All these results illuminate the potential of our method to make ASISR better performable in practical applications.

The remainder of this paper is organized as follows. Sec. 2 provides an overview of related work. Sec. 3 introduces the details of the proposed framework on Rot-E INR and ASISR architecture design, and also provides comprehensive equivariance error analysis on our method. Sec. 4 then demonstrates experimental results on evaluating the performance of the proposed method. The paper ends with a discussion of future work.



\section{Related Work}

\subsection{Arbitrary-scale Image Super-resolution}

Remarkable progress in deep learning-based image SR has been witnessed in the past decade \cite{wang2020deep, yang2019deep},
which can be majorly categorized to four model frameworks: Pre-Upsampling \cite{dong2014learning, kim2016accurate, shocher2018zero}, Post-Upsampling \cite{dong2016accelerating, lim2017enhanced, zhang2018image}, Progressive-Upsampling \cite{lai2017deep, lai2018fast, wang2018fully} and Iterative Up-and-Down Sampling \cite{haris2018deep, li2019feedback, fu2022kxnet}. Among these frameworks, the Post-Upsampling  framework
directly feeds the LR image as input of deep encoders, and the upsampling module 
is only adopted after the major part of the network. Post-Upsampling framework can thus significantly save computational cost, which make it received great research attention.  Most ASISR architectures also follow the Post-Upsampling SR framework, by replacing the traditional upsampling module with an INR-based module \cite{chen2021learning,lee2022local,song2023ope}. 

The early development of Post-Upsampling SR more focused on the design of the encoder for  feature extraction \cite{lim2017enhanced, zhang2018residual, zhang2018rcan, chen2021pre, liang2021swinir}.
Multiple CNN based feature extraction methods were first developed for this task using a new module or connection design, such as the residual block design \cite{lim2017enhanced}, deep residual channel attention \cite{zhang2018rcan} and the dense residual connection design \cite{zhang2018residual}. Very recently, inspired by the success of the self-attention mechanism, transformer-based SR architectures \cite{chen2021pre, liang2021swinir} surpassed CNN-based architectures using a larger data set and more parameters. Nevertheless, these approaches still ignore the super-resolution for an arbitrary scale factor and regard the super-resolution of different scale factors as independent tasks.

\red{ASISR is hopeful to greatly benefit practical utilization of SR models, since tracing and storing models individually for each specific scale factor would take evident memory resources, and testing with different models also causes additional time costs to re-perform feature extraction for different scale factors \cite{liu2024arbitrary, zhang2020multi, hu2019meta, chen2021learning,lee2022local,song2023ope,wang2021learning,son2021srwarp}.} This task is thus becoming one recent popular research topic in computer vision field. MetaSR \cite{hu2019meta} first proposed an ASISR model that dynamically predicts the weights of the up-scale filters under an arbitrary scale factor. Later, ArbSR \cite{wang2021learning}  constructed a general plug-in module using conditional convolutions, and conducted SR with different scales along the horizontal and vertical axes, respectively. In addition,
SRWarp \cite{son2021srwarp} adopted a differentiable adaptive warping layer to transform an LR image into HR representation.

As a more powerful ASISR framework, the LIIF was proposed \cite{chen2021learning}, firstly arising the concept of learning a continuous representation for the input LR image, and achieves better ASISR effect under this concept. Specifically, LIIF replaces the convolution-based upsampling module with an MLP-based INR module by taking continuous coordinates as input. Through training this INR module, the local implicit image function can then be obtained in a relatively simple but elegant manner. Following this groundbreaking work, to better understand the high-frequency components, LTE \cite{lee2022local} and UltraSR \cite{xu2021ultrasr} cleverly mapped the input feature and coordinates into a learnable high-dimensional Fourier space before passing them into the INR module. This mechanism makes the INR more suitable and interpretable for image representation and helps to achieve a significant performance improvement. To further reduce the complexity of the network in INR, OPE \cite{song2023ope} designed an orthogonal position encoding strategy, which is in a linear combination formulation without network layers. As a result, OPE not only makes the INR parameters free but also makes the INR more controllable.

To further enhance the ASISR performance, more carefully designed and complex modules have been attempted for the task nowadays. For example, CiaoSR \cite{cao2023ciaosr} constructed a continuous implicit attention network for better utilization of the local implicit image function and non-local information. SADN \cite{wu2021scale} further introduced a refined multi-scale feature extraction framework for ASISR network design. CLIT then \cite{chen2023cascaded} integrated the attention mechanism and frequency encoding technique into a unified local implicit image function. 
Besides, LINF \cite{yao2023local} modeled the distribution of texture details under different scaling factors with normalization flow. Furthermore, IDM \cite{gao2023implicit} integrated the diffusion framework \cite{ho2020denoising} for the ASSR task.
The ASISR technique has also been applied in more tasks such as depth super-resolution \cite{tang2021joint}, audio super-resolution \cite{kim2022learning}, and computed tomography \cite{zang2021intratomo}.

\red{
Although significant progress has been made, rotation equivariance has  not yet been explored for ASISR task. 
Under the highly ill-posed settings of ASISR, it is crucial to design a Rot-E framework for preserving the  original orientations and structural integrity of local geometric patterns, and enhancing the  reconstruction  stability and quality.
We thus aim to investigate the potential of Rot-E ASISR networks in this study.}

\subsection{Rotation Equivariant Deep Networks}
Early attempts for exploiting rotation symmetry prior in images were mainly designed by heuristics \cite{krizhevsky2012imagenet, laptev2016ti, esteves2017polar,sohn2012learning, he2015delving, Zhou2017Oriented, Marcos2017Rotation, Worrall2017Harmonic}.
Data augmentation \cite{krizhevsky2012imagenet} is the simplest and most commonly used among them, which enriches the training set with rotated samples to train a model with robustness to rotations.
This category of approaches suffers from the lack of theoretical guarantees and the unstable performance in keeping rotation equivariance.

Recent works have focused more on incorporating rotation equivariance directly into the network architecture. G-CNN \cite{Cohen2016group} first successfully constructed the framework of equivariant convolutions, which for the first time achieves a theoretically guaranteed Rot-E network for $\nicefrac{\pi}{2}$ rotations. Later, many Rot-E approaches have been proposed to enrich this framework \cite{hoogeboom2018hexaconv, weiler2018learning, weiler2019general, shen2020pdo, shen2021pdo, xie2022fourier}. Among these approaches, \cite{weiler2018learning} and \cite{weiler2019general} initially exploited the filter parametrization technique for arbitrarily rotating filters in the continuous domain with efficiency, and achieve arbitrarily degree rotation equivariance. \cite{shen2020pdo} and \cite{shen2021pdo} then proposed PDO-eConv by relating convolutions with partial differential operators, and first provided the error analysis. Besides, the Rot-E transformers have also achieved research interest and  been proposed for high-level version tasks \cite{he2021efficient, romero2020group, xu20232, hutchinson2021lietransformer}.  \red{In addition, theoretical researches has proven that equivariant models (such as group equivariant deep networks) achieve stronger generalization ability under appropriate tasks \cite{elesedy2022group,petrache2023approximation,sokolic2017generalization}.}

\red{For image restoration tasks, F-Conv \cite{xie2022fourier} exploited Fourier bases for high-accuracy filter parametrization and validated that Rot-E convolution can effectively help improve the performance of SR tasks. Then, \cite{celledoni2021equivariant} establishes  the relationship between the invariance of the regularization term and the equivariance of the proximal operator in inverse problems and provide the  principle for Rot-E proximal network design. 
\cite{fu2024rotation}  for the first time  provided theoretical analysis for Rot-E networks with arbitrary layers, and substantiated that rotation equivariant amelioration for unfolding networks consistently achieves state-of-the-art (SOTA) performance in multiple image restoration tasks, such as image super-resolution, CT image reconstruction, and image deraining.
\cite{terris2024equivariant} proposed an equivariant plug-and-play (PnP) image restoration method, validating that equivariance embedding can strongly improve the stability and reconstruction quality in PnP image restoration, like image deblur, image denoising, and MRI reconstruction.  Besides, rotation equivariant amelioration has also been validate to be effective in other image restoration tasks  like   Multi-modality image funsion \cite{zhao2024equivariant} and  hyperspectral image inpainting\cite{li2024equivariant} in  recent years.}

\red{Although Rot-E modules have been introduced to enhance various network architectures for different tasks, designing an ASISR framework that ensures end-to-end rotational equivariance throughout the entire network flow remains a significant challenge.  Particularly, Rot-E INR for ASISR is still unexplored and the representation accuracy of Rot-E encoder can be further enhanced.
This study is dedicated to solving these issues.}

\section{Proposed Rot-E Framework for ASISR}
The proposed ASISR framework consists of two parts,  Rot-E encoder (as shown in Fig. \ref{fig:method}(a)) and  Rot-E INR (as shown in Fig. \ref{fig:method}(b)).
In this section, we first introduce some necessary preliminary knowledge on Rot-E network structure.
Then, we introduce the proposed Rot-E modules for INR, as well as the theoretical rotation analysis for both the proposed individual modules and entire network constructed with the proposed framework. Finally, we present the proposed enhancement for the Rot-E encoder.

\subsection{Preliminary Knowledge on Rot-E Structure}
The G-CNN based network is the most commonly utilized Rot-E architecture \cite{weiler2019general,  shen2020pdo, xie2022fourier}, whose capability has been widely validated by both comprehensive experiments and solid theories. Specifically, the encoder module of the expected Rot-E ASISR can be easily designed by directly employing this type of architecture. We first introduce some necessary knowledge about such a Rot-E structure.

\textbf{Definition of rotation equivariance.}
Following the definition in previous works, rotation equivariance means that a rotation operator imposed on the input will result in rotation on the output, without causing any other unpredictable changes.
Mathematically,  let $\Psi$ be a mapping from the input feature space to the output feature space, and $S$ be a group of rotation transformations, i.e., \begin{equation}\label{Group}
S=\l\{A_k =
\begin{bmatrix}
    \cos \nicefrac{2\pi k}{t} &  \sin \nicefrac{2\pi k}{t} \\
   -\sin \nicefrac{2\pi k}{t} &  \cos \nicefrac{2\pi k}{t}
  \end{bmatrix}\mid k = 0,2,\cdots,t-1
\r\}.
\end{equation}
Then, $\Phi$ is equivariant with respect to $S$ if for any rotation matrix ${\tilde{A}}\in S$, it holds that
\begin{equation}\label{equivariance}
   \Phi\[ \pi_{\tilde{A}}^I\](I)  = \pi_{\tilde{A}}^F\[\Phi\](I),
\end{equation}
where $I$ represents the input image;  $\pi_{\tilde{A}}^I$ and $\pi_{\tilde{A}}^F$ denote how the transformation ${\tilde{A}}$ acts on input image and output features, respectively;  $[~\!\cdot\!~]$ denotes the composition of functions.

\textbf{Structure of feature map.}
For an input image $I$ of size $h\times w \times n_0$ ($n_0=1$ for gray images and $n_0=3$ for color ones), we denote the intermediate feature map achieved by the Rot-E network as $F$.
As shown in Fig. \ref{fig:eq_orl}, $F$ corresponds to a multi-channel  tensor of size $h\times w \times n \times t$,  where the third dimension is with respect to the feature channel, and the fourth dimension is with respect to a chosen rotation group $S$.
Following the previous works \cite{shen2020pdo, xie2022fourier},  we denote the feature map corresponding to the $k$-th group element
as $F^{A_k}\in\R^{h\times w \times n}$,  where $A_k$ is a rotation matrix in $S$, and also used as an index for denoting a specific tensor mode in $F$.

\textbf{Rotation on feature map.} In Rot-E networks \cite{weiler2018learning, shen2021pdo, xie2022fourier}, the rotation of the input image will cause rotation of the feature map. However, there is an additional mode with respect to the rotation group in $S$, which makes the rotation of  feature map more complex than the rotation of image. As shown in Fig. \ref{fig:eq_orl}, the rotation of feature map would be consistent with spatial rotation on the first two dimensions and cyclic shifting along the final dimension. Formally, for any ${\tilde{A}}\in S$, we have
\begin{equation}\label{rotation}
  \pi_{\tilde{A}}^F(F) \!=\! \[\pi_{\tilde{A}}^I\!\(F^{{\tilde{A}}^{\!-\!1}A_1}\!\)\!, \!\pi_{\tilde{A}}^I\!\(F^{{\tilde{A}}^{\!-\!1}A_2}\!\)\!,\! \cdots\!,\!\pi_{\tilde{A}}^I\!\(F^{{\tilde{A}}^{\!-\!1}A_t}\!\) \]\!,
\end{equation}
where $F^{A_k}, k =1,2,..,t$  are tensors of size $h\times w\times n$, which can be viewed as an $n$-channel image when performing spatial rotation $\pi^I_{\tilde{A}}$.
One can refer to the supplementary material for more details about the rotation operators on image and feature map in Rot-E networks.

\subsection{Rotation Equivariant INR Framework}
To construct an end-to-end rotation equivarent network for AISSR, realizing a rotation equivariant INR module is inevitable. This challenging task represents the primary focus of this study. To this aim, we first clarify the rotation equivariance essence on INR as follows. 

\subsubsection{Rotation Equivariance on INR}
As shown in Fig. \ref{fig:eq_ASSR}, a local implicit image function corresponds only to one single pixel on the feature map, while a pixel in the feature map only corresponds to the image within the receptive field of its position. We thus only need to consider how the rotation of the image within a receptive field affects the local image function at each position.

Denote the feature tensor at pixel $(i,j)$ as $F_{ij}$ (which is of size ${n\times t}$,  stacked by  vectors $F^{A_k}_{ij}\in\mathbb{R}^{n}$,  $k = 1,2,\cdots,t$).
It is easy to deduce that, following a Rot-E encoder, applying rotation $A\in S$ to the image within the receptive field at position $(i,j)$  would only cause cyclic shifting on the final dimension of $F_{ij}$. Then, the key issue for constructing the Rot-E ASISR network is to consider an INR whose input coordinates can rotate in correspondence with the cyclic shifting on $F_{ij}$.

\begin{figure}
    \centering
    \includegraphics[width=1\linewidth]{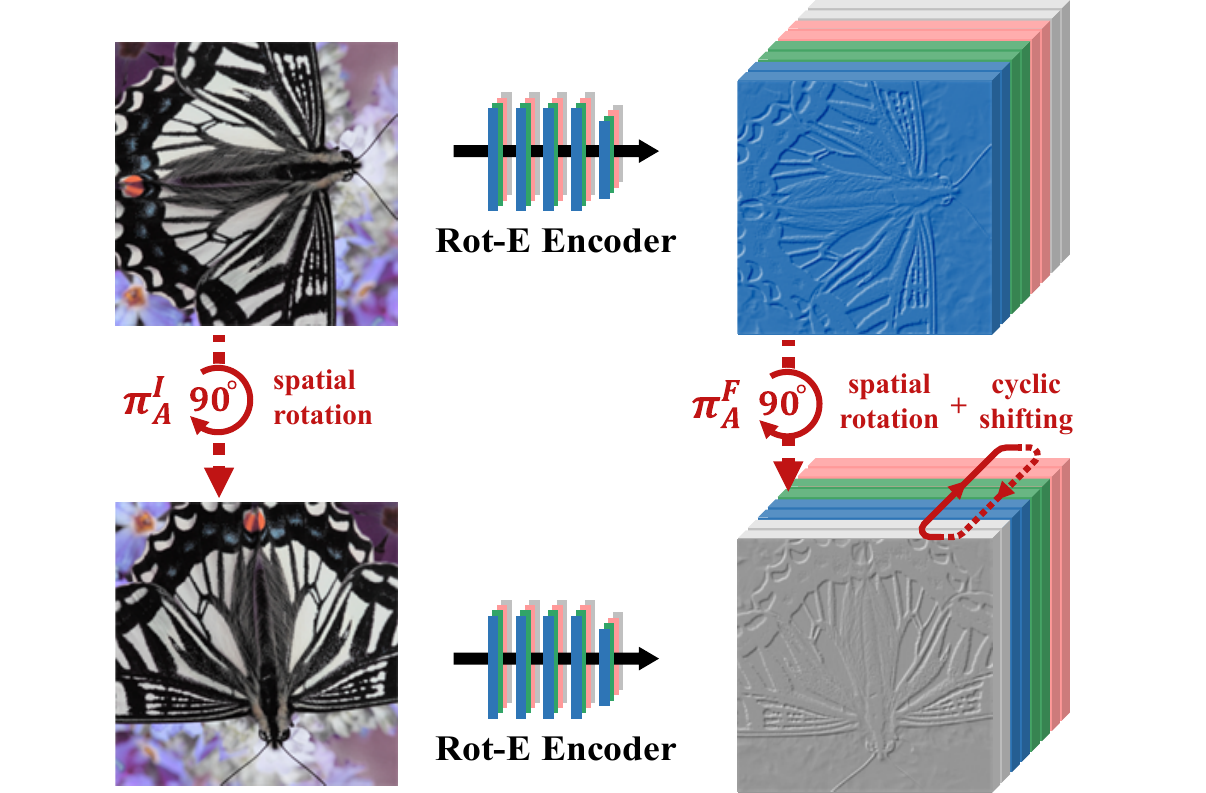}
    \caption{Illustration of the rotation operator imposed on the input and its corresponding feature map of a Rot-E network.}
    \label{fig:eq_orl}
\end{figure}

Formally, we can conclude that the local operation of a Rot-E INR, denoted as $\INR$, should satisfy:
\begin{equation}\label{EQonLocal}
\INR\(\pi^F_{\tilde{A}}\(F_{ij}\)\) = \pi^f_{\tilde{A}}\[ \INR\(F_{ij}\)\],
\end{equation}
where for an arbitrary 2D function $f(x)$,
\begin{equation}\label{rotf}
\pi^f_{\tilde{A}}[f](x) = f\(\tilde{A}^{-1}x\)
\end{equation}
represents the coordinate rotation of 2D functions.
Besides, by substituting $F_{ij}$ into Eq. \eqref{rotation}, we can obtain
\begin{equation}\label{rotF}
  \pi^F_{\tilde{A}}(F_{ij}) =   \[F^{{\tilde{A}}^{-1}A_1}_{ij}, F^{{\tilde{A}}^{-1}A_2}_{ij}, \cdots,F^{{\tilde{A}}^{-1}A_t}_{ij} \],
\end{equation}
which is indeed the cyclic shifting on $F_{ij}$. One can see Figs. \ref{fig:eq_orl} and \ref{fig:eq_ASSR} to more intuitively understand this kind of rotation equivariance and know its specific characteristic by comparing it with previous Rot-E networks.

\begin{figure}
    \centering
    \includegraphics[width=1\linewidth]{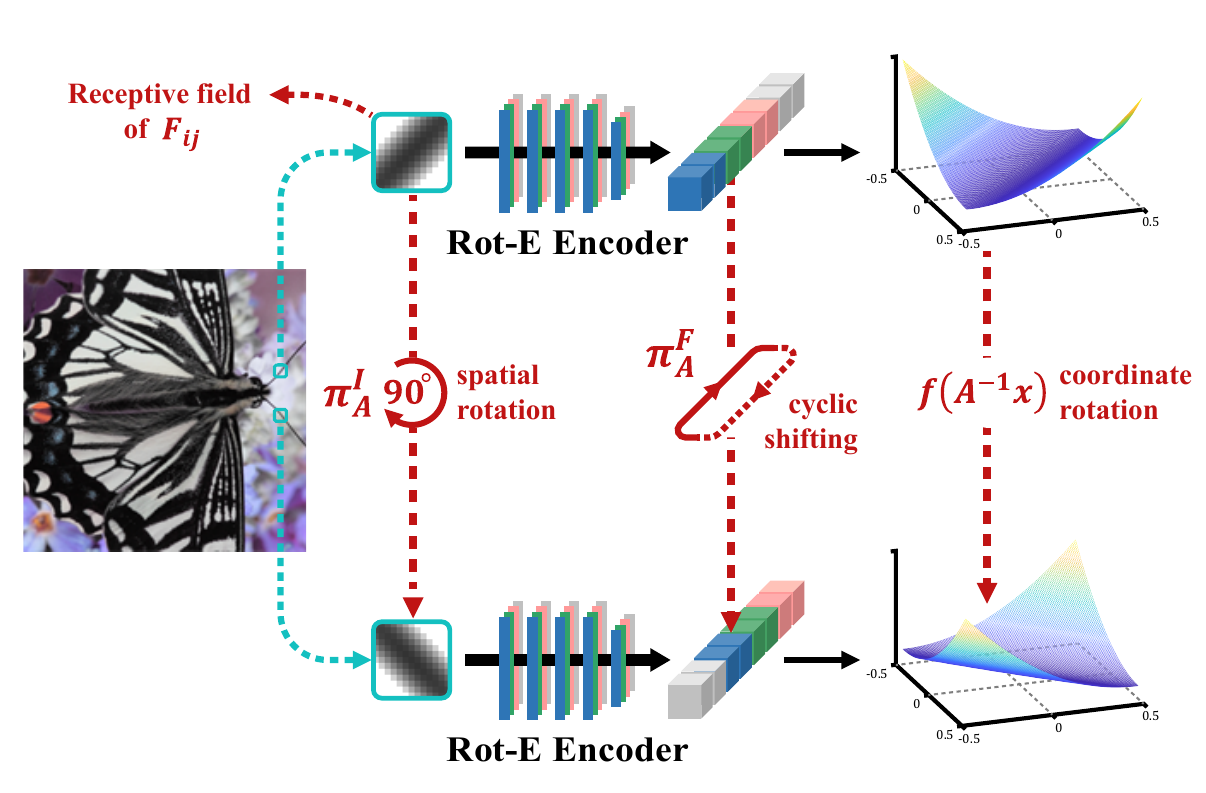}
    \caption{Illustration of the rotation equivariance in ASISR tasks and the rotation on local implicit image function.}
    \label{fig:eq_ASSR}
\end{figure}

\begin{figure*}
    \centering
    \includegraphics[width=\linewidth]{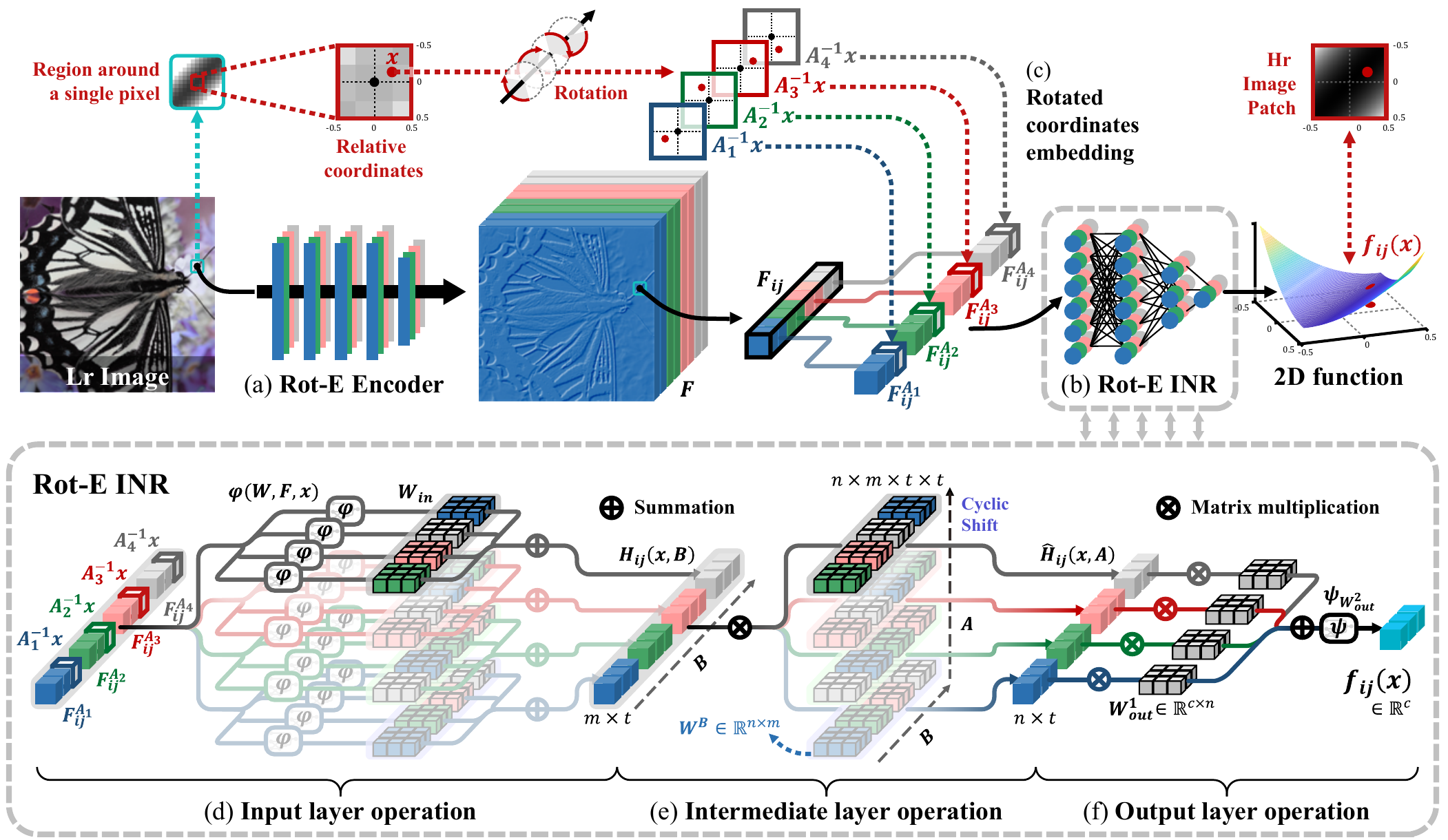}
    \caption{Illustration of the framework of the proposed Rot-E ASISR; (a) the Rot-E encoder, constructed with Rot-E convolutions or Rot-E transformers; (b) the Rot-E INR module; (c) lifting the coordinate by rotating it with all elements of the rotation subgroup; (d)-(f) the designs of the input layer, intermediate layer and output layer, respectively. }
    \label{fig:method}
\end{figure*}

\subsubsection{Rotation Equivariant INR Module Design}

For constructing a Rot-E INR satisfying the above condition, we need to carefully design new INR modules, including the input layer, intermediate layer, and output layer, respectively.

\textbf{Input layer for Rot-E INR.}
As shown in Fig. \ref{fig:method}(b) and (d), the proposed input layer of Rot-E INR takes each pixel of the feature map $F$ (provided by Rot-Encoder) as input.
Besides, the output of the input layer of Rot-E INR is a set of functions. Here,  we denote it as $H_{ij}(x, B)$ for any $B\in S$, while $x$ is the high-resolution coordinate.
One can view Fig. \ref{fig:method} for easy understanding of the structure of $F$ and $H_{ij}(x, B)$, respectively.

Unlike traditional ASISRs shown in Fig. \ref{fig:fig1}, which directly  concatenates  $x$  with $F_{ij}$ and input them  into a selected function to construct an INR, we first lift the coordinate $x$ by rotating it with all the elements of the rotation subgroup before utilizing it.  As shown in Fig. \ref{fig:method}(c), the rotated coordinates are then concatenated with the feature map associated with the corresponding group elements and then fed into a selected function.
Formally,  the operation of the input layer is denoted as $H_{ij}(x, B) = [W_{\inp}\hat{\circ}_\varphi F_{ij}](x,B)$, which can be calculated by
\begin{equation}\label{input_layer}
\[W_{\inp}\hat{\circ}_\varphi F_{ij}\](x,B)=\sum_{A\in S}\varphi\(W_{\inp}^{B^{-1}A}, F^A_{ij}, A^{-1}x\),
\end{equation}
where  $\varphi$ is an arbitrary function selected for constructing INR, and
 $W_{\inp}=\{W_{\inp}^{A}\mid A \in S\}$ denotes the possible learnable parameters in this layer,  which can be divided into $t$ similar parts, corresponding to the $t$ elements in $S$.  As shown in Fig. \ref{fig:method}(d), the elements of $W_{\inp}$ are shifted cyclically along the group index $A$, just as the convolution filters are shifted cyclically
along the group index in the G-CNN framework \cite{weiler2019general}.
The equivariance principle of this design will be demonstrated in the following section.

It should be noted that $\varphi$ can be an arbitrary function chosen for constructing INR. This is crucial for ensuring that the proposed framework can transform various kinds of INR methods into their Rot-E versions in a PnP manner.
we will provide typical examples in Sec. 3.2.4 to help more easily
understand how to set \eqref{input_layer} in practice.

\textbf{Intermediate layer for Rot-E INR.}
The intermediate layer of Rot-E INR maps a group of functions $H_{ij}(x,B)$ to another group of functions $\hat{H}_{ij}(x,A)$.
Specifically, for any $x,A,B$, we have $H_{ij}(x,B)\in\mathbb{R}^m$ and $\hat{H}_{ij}(x,A)\in\mathbb{R}^n$,  where $m$ and $n$ are the channel numbers of $H(x,B)$ and $\hat{H}(x,A)$, respectively. Then, the operation of the intermediate layer can be denoted as $\hat{H}_{ij}(x,A) = [W\tilde{\circ}H_{ij}](x,A)$, which is calculated by
\begin{equation}\label{inter_layer}
  \[W\tilde{\circ}H_{ij}\](x,A) = \sum_{B\in S} W^{A^{-1}B}\cdot H_{ij}(x,B),
\end{equation}
where $W$ represents the learnable parameters,  which consist of $W^A\in\mathbb{R}^{n\times m}$, $A\in S$, and ``$\cdot$'' denotes the matrix multiplication. Note that $W$ is cyclically shifted along the group index just as the calculation in Eq. \eqref{input_layer}. In the following section,  we will show that Eq. \eqref{inter_layer} is a rotation equivariant linear layer \cite{ravanbakhsh2020universal}.

\textbf{Output layer for Rot-E INR.}
The output layer of the Rot-E INR maps a set of functions $\hat{H}_{ij}(x,A)$ to a 2D function $f_{ij}(x)$. Specifically, we denote it as $f_{ij}(x) = [W_{\outp}\check{\circ}\hat{H}_{ij}](x)$, which is calculated by
\begin{equation}\label{output_layer}
  \[W_{\outp}\check{\circ}_\psi\hat{H}_{ij}\](x) = \psi_{W^2_{\outpp}}\(\sum_{A\in S} {W^1_{\outp}}\cdot \hat{H}_{ij}(x,A)\),
\end{equation}
where $W_{\outp}=\l\{W_{\outp}^1, W_{\outp}^2\r\}$, and $W_{\outp}^1\in \mathbb{R}^{m\times n}$ is the learnable matrix, with $m$ being the selected feature channel number, $\psi_{W_{\outpp}^2}$ denotes an arbitrary function with $W_{\outp}^2$ being its learnable parameters.

In practice, $\psi_{W_{\outpp}^2}$ can be set as arbitrary functions  (e.g., deep networks like MLP), which further enhances the design flexibility. In particular, when $\psi_{W_{\outpp}^2}$ is set as an identity function and let $m=n_0$, the output layer becomes a commonly used linear layer. 


\begin{figure*}
    \centering
    \includegraphics[width=\linewidth]{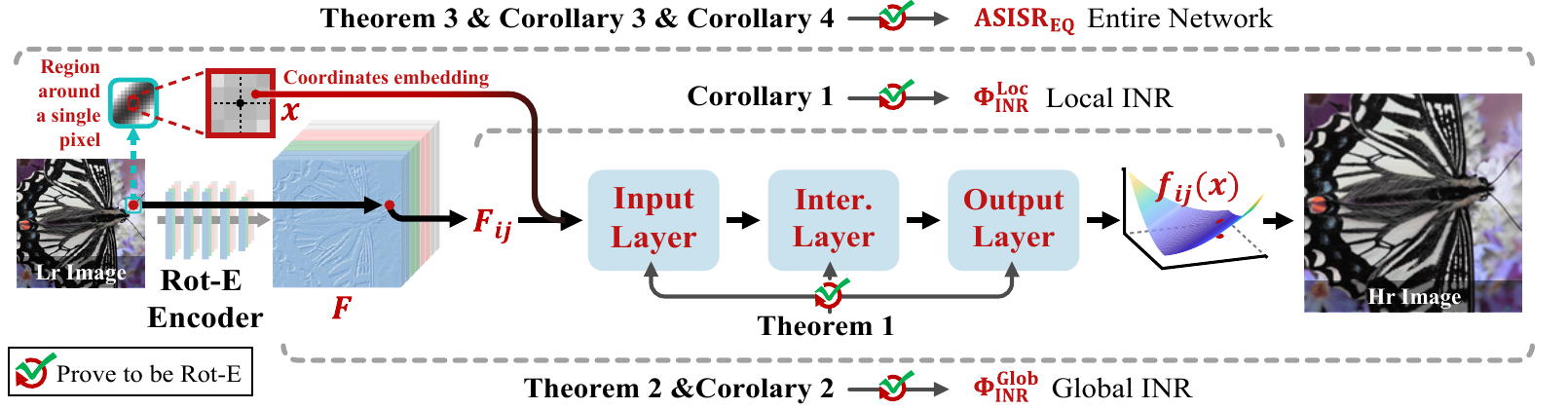}\vspace{-3mm}
    \caption{\red{Illustration of the proposed theoretical conclusions, which shows the correspondence between each theoretical conclusion and its respective network module.} }
    \label{fig:Theorem}
\end{figure*}

\textbf{Constructing Rot-E INR.}
We can now readily construct Rot-E INRs for local implicit image function representation, by connecting one input layer, several intermediate layers (optional), and one output layer together.  Adding any pixel-wise function (such as the ReLU function) and other equivariant modules (e.g., equivariant batch normalization and equivariant dropout \cite{weiler2018learning, weiler2019general}) to any position of the network will not affect the overall equivariance.
Once the local implicit image function is achieved, we can further obtain the entire continuous SR image by assembling all the local implicit functions at each pixel of the LR image.

Formally,  we denote the local operation and the global operation of the INR constructed by the proposed modules as $\INR$ and $\INRG$, respectively. Specifically, 
 $\INR$ maps an input feature map to a local implicit image function, i.e., $f_{ij}= \INR(F_{ij})$. It can be represented in the following formulation:
\begin{equation}\label{local}
\INR(F_{ij}) =  W_{\outp}\check{\circ}_\psi W_L\tilde{\circ}\!\cdots\!\tilde{\circ}W_2\tilde{\circ} W_1\tilde{\circ}W_{\inp}\hat{\circ}_\varphi F_{ij},
\end{equation}
where $L$ is the number of intermediate layers. It should be noted that we neglect the activation functions in Eq. \eqref{local} for the conciseness of the expression, which could be added to any position of the network without affecting its rotation equviariance.
Besides, the entire image can be achieved by stacking the loccal INRs, i.e.,  
 for any $x\in\mathbb{R}^2$, the global continuous image $\SR(x)$ can be calculated by
\begin{equation}\label{SR}
  \SR(x) = \INRG(F)(x) = \INR(F_{\hat i\hat j})(x-x_{\hat i\hat j}),
\end{equation}
where $x_{\hat i\hat j}$ denotes the  coordinate of the pixel at position $(\hat i, \hat j)$ in the LR image, $(\hat i, \hat j)$ is the position of the pixel nearest to $x$ in the LR image, i.e.,  $(\hat i, \hat j)=\arg\min_{(i,j)}\| x - x_{ij}\|$.
Combining this module with the Rot-E encoder $\Encoder$, a whole Rot-E ASISR can be represented as:
\begin{equation}\label{ASISRall}
   \ASSR = \INRG\[\Encoder\],
\end{equation}


\subsubsection{Equivariance Analysis for the proposed Rot-E INR}

We will then prove that the ASISR constructed with the proposed modules satisfies the rotation equivariance as shown in Fig. \ref{fig:eq_ASSR}.

In order to rigorously show the rotation equivariance of the proposed INR module,
we first provide the mathematical definition of the rotation of the feature functions $H_{ij}(x,A)$.
\red{Similar to the rotation of feature layers of group convolutions (as shown in the right part of Fig. \ref{fig:eq_orl}), 
The rotation here consists of the coordinate rotation in the spatial dimension ($x\to \tilde{A}^{-1}x$) and the cyclic shifting in the group dimension ($A\to \tilde{A}^{-1}A$). Mathematically,  
for  any $x\in\mathbb{R}^2$ and $A\in S$,  the rotation on $H_{ij}(x,A)$ corresponding to  any $\tilde{A} \in S$ is defined as}:
\begin{equation}\label{rotH}
\begin{split}
  &\pi^H_{\tilde{A}}[H_{ij}](x,A) = H_{ij}\({\tilde{A}}^{-1}x,{\tilde{A}}^{-1}A\),\\
\end{split}
\end{equation}
where $[~\cdot~]$ denotes the composition of functions.

\red{Then, we can formally analyze the rotation equivariance of the proposed INR. Specifically, we can achieve the following theoretical results.
Please refer to Fig. \ref{fig:Theorem} for clear correspondence between the theoretical results and their respective network modules' equivariance properties.
Besides, 
please see more intuitive and detailed illustration of the theoretical results and  proof details in the supplementary material. }

\textbf{Equivariance of INR on local implicit image function.}  \red{We first prove that the proposed three INR modules are indeed rotation equivariant, i.e.,  rotation on the input of each INR module induces a corresponding rotation on the output of the module.}  Specifically, we have following theorem:
\begin{Thm}\label{ThmLocal}
For any pixel position $(i,j)$,  let $F_{ij}$ denote the feature provided by a Rot-E encoder with its equivariant group $S$;  let  $W_{inp}$, $W$ and $W_{\outp}$ be the parameters defined in \eqref{input_layer}, \eqref{inter_layer} and \eqref{output_layer}, respectively; and let $H_{ij}$ denote the feature function achieved  by the proposed input layer or intermediate layer,
 respectively, then the following results are satisfied for  $\forall \tilde{A}\in S$:
\begin{equation}\label{thm1}
\begin{split}
  W_{\inp}\inpC \(\pi^F_{\tilde{A}}\(F_{ij}\)\) &= \pi^H_{\tilde{A}}\[W_{\inp}\inpC  F_{ij}\],\\
  W \intC \(\pi^H_{\tilde{A}}\[H_{ij}\]\) &= \pi^H_{\tilde{A}}\[W\intC H_{ij}\],\\
  W_{\outp}\outC\(\pi^H_{\tilde{A}}\[H_{ij}\]\) &= \pi^f_{\tilde{A}}\[W_{\outp}\outC H_{ij}\],\\
\end{split}
\end{equation}
\red{where  $\pi_{\tilde{A}}^F$, $\pi_{\tilde{A}}^H$ and $\pi_{\tilde{A}}^f$
are rotation operators on $F$, $H$ and output function respectively, which are defined in \eqref{rotF} and  \eqref{rotH}}.
\end{Thm}

Theorem \ref{ThmLocal} shows that each of the proposed modules is theoretically rotation equivariant with the rotation defined in \eqref{rotF} and \eqref{rotH}.
Based on the above theorem, it is easy to deduce that any INR constructed with the proposed modules is Rot-E on local implicit image function, i.e.,  we have the following corollary to Theorem \ref{ThmLocal}:

\begin{Cor}\label{CorLocal}
For any INR constructed by the proposed modules, denote its local operation as $\INR$. Then, for arbitrary pixel position $(i,j)$, let $S$ denotes the equivariant group, and the following result is satisfied for any $\tilde{A}\in S$:
\begin{equation}\label{cor1}
\INR\(\pi^F_{\tilde{A}}\(F_{ij}\)\) = \pi^f_{\tilde{A}}\[ \INR\(  F_{ij}\)\].
\end{equation}
\end{Cor}

\red{The corollary formally shows that
the implicit image function constructed with the proposed INR framework can coordinately rotate with respect to the cyclic shifting on $F_{ij}$. In other words, it’s rotation equivariant.}

\textbf{Equivariance of INR on the entire HR images.} \red{The entire HR image is achieved by stacking the loccal INRs, and therefore, based on  Corollary \ref{CorLocal}, we can provide the rotation equivariance analysis on the entire image.} Formally, we have the following results:

\begin{Thm} \label{INRG}
 Assume that the output feature map of a Rot-E encoder, denoted as $F$, is discredited from a smooth function $e:\R^2\times S\to\R$, whose channel number is $n$,  and the channel number of the SR image is $n_0$. Denote the equivariant group as $S$ with  $|S| = t$,  and denote  the global operator corresponding to the INR constructed with the proposed modules  as  $\INRG$. Then,
if the following conditions are satisfied for any $ x\in\mathbb{R}^2$, $ A \in S$, $c = 1,2,\cdots, \frac{n}{t}$ and $c_0 = 1,2,\cdots, n_0$:
\begin{equation}\label{con1}\
\begin{split}
&\l\|\nabla_x e_{c}(x,A)\r\|\leq G_e, \\
&\l\|\nabla_x \(\INRG(F)(x)\)_{c_0} \r\|\leq G_{\emph{INR}}^x,  \\
&\sup_{i,j}\l\{\l\|\nabla_{F_{ij}} \(\INRG(F)(x)\)_{c_0} \r\|\r\}\leq G_{\emph{INR}}^F,
\end{split}
\end{equation}
then for any $\tilde{A}\in S$ and $x\in \mathbb{R}^2$, the following result is satisfied\footnote{For a tensor $X$, $|X|\leq \epsilon$ means that the absolute value of all the elements in $X$ are not greater than $\epsilon$.}:
\begin{equation}\label{SRerror}
  \l|  \INRG\(\pi_{\tilde{A}}^F(F)\)(x) -   \pi_{\tilde{A}}^f\[\INRG\(F\)\] (x) \r| \leq C\delta,
\end{equation}
where $C=\sqrt{2n}G_{\emph{INR}}^F G_e  + \sqrt{2}G_{\emph{INR}}^x$ and $\delta$ is the mesh size of the LR image.
\end{Thm}

Theorem \ref{INRG} shows that the rotation equivariance error of the proposed INR mainly depends  on the mesh size and the partial derivative of INR. When the mesh size of the LR image approaches zero, the equivariant error will also approach zero, similar to the rotation equivariant CNNs designed in previous research \cite{weiler2018learning, shen2021pdo, xie2022fourier}. Besides, it is easy to find that the partial derivatives $ G_{\emph{INR}}^x$ and $G_{\emph{INR}}^F$ are usually bounded numbers in common INRs.

Moreover for most commonly used rotation groups (i.e., p4 group, i.e., $\frac{\pi}{2}$  rotations), we can prove that the rotation equivariance error of the proposed INR is zero. Formally, there is the following corollary for Theorem \ref{INRG}.

\begin{Cor}\label{CorG}
Under the same condition of Theorem \ref{INRG}, when $t = 2$ or $t =4$, the following result hold: for any $\tilde{A}\in S$:
\begin{equation}\label{Cor2}
\INRG\(\pi_{\tilde{A}}^F(F)\) =\pi_{\tilde{A}}^f\[\INRG\(F\)\].
\end{equation}
\end{Cor}

\textbf{Equivariance of the entire Network.} Finally, we analyze the Rot-E of the entire network constructed by combining commonly used Rot-E-convolution-based encoder \cite{weiler2018learning, shen2021pdo, xie2022fourier} with the proposed Rot-E INR. Specifically, we can deduce the following results.

\begin{Thm} \label{EntireNet}
For an image ${I}$ with size $h\times w\times n_0$, and an ASISR method constructed by $\ASSR = \INRG\[\CNN\]$, where $\CNN$ is a Rot-E encoder constructed by $L$-layer rotation equivariant CNN network, whose channel number of the $l^{th}$ layer is $n_l$, rotation equivariant subgroup is $S\leqslant O(2)$, $|S|=t$, and activation function is set as ReLU. $\INRG$
denotes the global operator corresponding to the INR constructed with the proposed modules.
Denote the latent continuous function of the $c^{th}$ channel of ${I}$  as $r_c: \mathbb{R}^2 \! \rightarrow \! \mathbb{R}$, and the latent continuous function of any convolution filters in the $l^{th}$ layer  as $\phi^{l}: \mathbb{R}^2 \! \rightarrow \! \mathbb{R}$. Assuming for any $x\in \R^2$, $l \in \{1, \cdots, L\}$, $c \in \{1, \cdots, n_{0}\}$,  and any $F$ produced with Rot-E encoder, $ A \in S$,  the following conditions are satisfied:
\begin{equation}
        \begin{split}
            & \l\|\nabla_x \(\INRG(F)(x)\)_{c} \r\|\leq G_{\emph{INR}}^x, \\
            & \sup_{i,j}\l\{\l\|\nabla_{F_{ij}} \(\INRG(F)(x) \)_{c}\r\|\r\}\leq G_{\emph{INR}}^F,\\
            & |r_c(x)| \leq F_0, \|\nabla_x r_c(x)\| \leq G_0, \|\nabla_x ^2 r_c(x)\| \leq H_0, \\
            & |\phi^{l}(x)| \leq F_l, \|\nabla_x \phi^{l}(x)\| \leq G_l, \|\nabla_x ^2 \phi^{l}(x)\| \leq H_l,\\
            &\forall \|x\|\geq\nicefrac{(p+1)\delta}{2},~ \varphi_{l}(x)=0,\\
        \end{split}
\end{equation}
where $p$ is the filter size, $\delta$ is the mesh size, and $\nabla$ and ${\nabla}^2$ denote the operators of gradient and Hessian matrix, respectively.
Then, for   any $\tilde{A}\in S$ and $x\in\mathbb{R}^2$, the following result is satisfied:
\begin{equation}\label{SRerror}
  \l|  \ASSR\!\(\pi_{\tilde{A}}^I(I)\)\!(x) \!-\!   \pi_{\tilde{A}}^f\[\ASSR\(I\)\]\!(x)   \r|\! \leq\! C \delta,
\end{equation}
where $\delta$ is the mesh size of the LR image,
\begin{equation}\label{C1C2}
  \begin{split}
     \!\!C\!=\!&\(\!\!\sqrt{2n_L}G_{\emph{INR}}^F\!\!\(\sum_{m=1}^{L} \!  \frac{G_mF_0}{F_m}  \! +\! G_0\!\)\!  \mathcal{F} \!\! +\! \!\sqrt{2}G_{\emph{INR}}^x\!\)\!\!+\!O(\delta),
  \end{split}
\end{equation}
and $\mathcal{F} = \prod_{l=1}^L{n_{l-1}p^2F_l}$.
\end{Thm}

Theorem \ref{EntireNet} shows that the equivariant error of the entire network will also approach zero when the mesh size of the LR image approaches zero. It should be noted that the new element in Theorem \ref{EntireNet},  $\mathcal{F}=\prod_{l=1}^L{n_{l-1}p^2F_l}$,  has been  indicated  to be an estimable finite number in \cite{fu2023rotation}. This implies that $C$ is a finite constant.

For the most commonly used p4 rotation group, it is also easy to deduce that the equivariant error of the proposed ASISR is zero, i.e., we have the following corollary:

\begin{Cor}
Under the same condition of Theorem \ref{EntireNet},
and when $t = 2$ or $t =4$, the following result is satisfied for any $\tilde{A}\in S$:
\begin{equation}\label{Cor32}
     \ASSR\(\pi_{\tilde{A}}^I(I)\) =  \pi_{\tilde{A}}^f\[\ASSR\(I\)\].
\end{equation}
\end{Cor}

Moreover, we can further deduce that the bound of equivariant error of arbitrary rotation degrees (note that its rotations in the aforementioned theorem are all limited in finite rotation degrees corresponding to $S$), i.e., we have:

\begin{Cor}
Under the same condition of Theorem \ref{EntireNet}, for  an arbitrary $\theta \in \[0, 2 \pi \]$, let $A_{\theta}$ denote the $\theta$-degree rotation matrix, then
 for any $ \theta$ we have
\begin{equation}\label{Cor32}
    \l|  \ASSR\!\(\pi_{A_\theta}^I\!(I)\)\!(x) \!-\!   \pi_{A_\theta}^f\!\[\ASSR\(I\)\]\!(x)   \r| \!\leq\! C\delta+\hat{C}t^{-\!1}\!,
\end{equation}
where $\hat{C} = 4\pi n_LG_{\emph{INR}}^F\(\max\{h, w\}\!+\!L(p\!+\!1)\)$.
\end{Cor}

This corollary shows that the equivariant error is primarily influenced by two factors: the mesh size
$\delta$ and the equivariant orientation number $t$.

\subsubsection{Typical examples for Rot-E INR}

In this section, we will present typical examples that demonstrate how to apply the proposed modules to reform the INRs in current ASISR frameworks into their Rot-E forms.

\textbf{Rot-E LIIF.}   The most classical INR for ASISR is  LIIF \cite{chen2021learning}, which  takes a MLP  to map the coordinate of HR pixel and  the corresponding latent codes and  into image value, i.e., for the position $(i,j)$,
\begin{equation}\label{LIIF}
f^{\emph{LIIF}}_{ij}(x)= \MLP_M\(\begin{bmatrix}
                             F_{ij} \\
                             x
                           \end{bmatrix}\),
\end{equation}
where $M$ is the number of linear layers in MLP.
In order to achieve Rot-E LIIF, we should first replace the encoder into a correlated Rot-E encoder, and then
transform all the involved linear layers of LIIF into its Rot-E versions. 
 The general framework of Rot-E INR is in the formulation of \eqref{local}. Specifically, for the input layer \eqref{input_layer}, we
set  $\varphi$  as
\begin{equation}\label{Exam_1}
  \varphi(W, F, x) = W\cdot\begin{bmatrix}
                             F \\
                             x
                           \end{bmatrix},
\end{equation}
where $F$ is the involved feature vector,  $W$ is a learnable matrix.
For the output layer, we can set  the involved $\psi$ as a  MLP, i.e.,
\begin{equation}\label{Exam_1_out}
\psi\(F\) = \MLP_N\(F\),
\end{equation}
where $F$ is involved feature vector, $N = M-L-2$,  $L$ is the number of intermediate layers defined in \eqref{local}. It is easy to find that, in these settings, \eqref{local} actually become a MLP with appropriate parameter sharing in each layer, and the total layer number\footnote{one linear layer in input layer, $L$ linear layers in intermediate layers and $M-L-1$ linear layers in output layers.} is the same as the MLP in \eqref{LIIF}.

Since  the proposed intermediate layer  contains only $\frac{1}{t}$ of the parameters of generally used linear layer, when there is the same channel number (i.e., the same memory and computational cost in terms of FLOPs), we can control the quantity of parameters by changing the number of intermediated layers  $L$, i.e., larger $L$ will lead to fewer parameter in the INR.
One can simply set $L= 0$, in which case $\psi$ becomes an ($M-2$)-layer MLP, and  the constructed Rot-E LIIF is closest to the original LIIF. In this case, the Rot-E LIIF can be represented as:
\begin{equation}\label{EQ_LIIF}
  f_{ij}^{\emph{E-OPE}}(x)\!=\!\MLP_{N} \!\(\sum_{A,B\in S}W^1_{\outp}\! \cdot\! W_{\inp}^{B^{-1}A}\!\cdot\! \begin{bmatrix}
                             F^A_{ij} \\
                             A^{-1}x
                           \end{bmatrix}\),
\end{equation}
where $ W_{\inp}$ and $W^1_{\outp}$ are defined in \eqref{input_layer} and \eqref{output_layer}, respectively. 
It should be noted that the feature map $F$ provided by the Rot-E encoder is now with an additional channel with respect to the group index $A$, which is different from the feature map as expressed in Eq. \eqref{LIIF}.

\begin{figure*}
    \centering
    \includegraphics[width=\linewidth]{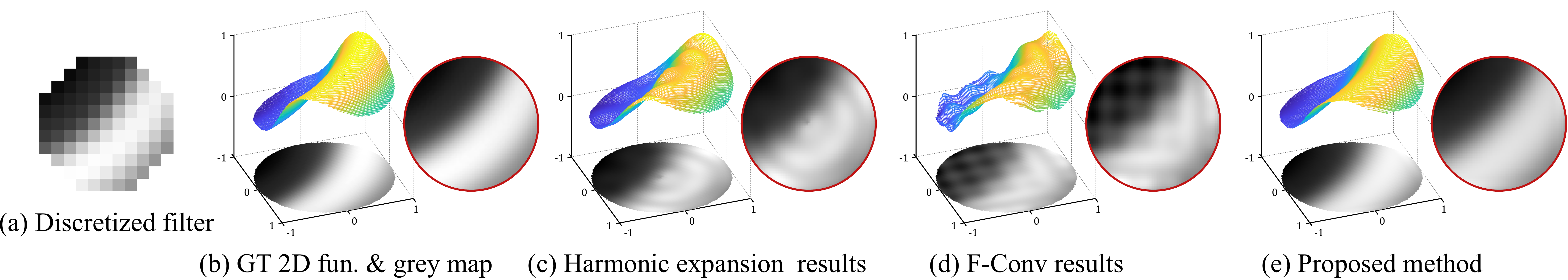}
    \caption{Illustration of the implicit function of a $11\times 11$ filter, achieved by different filter parametrization methods. (a) the $11\times 11$ discretized filter; (b) the ground truth 2D function and grey map; (c)-(e) the 2D function and grey map achieved by  Harmonic expansion (E2-CNN \cite{weiler2018learning, weiler2019general}), F-Conv \cite{xie2022fourier} and the proposed method, respectively.}
    \label{fig:bicubic}
\end{figure*}

\textbf{Rot-E OPE.} The orthogonal position
encoding (OPE) method is a parameter-free INR approach \cite{song2023ope}. It performs a linear combination of orthogonal 2D-Fourier basis functions, using the elements in the feature vector of the encoder as the combination coefficients. 
Formally, it can be calculated by
\begin{equation}\label{OPE}
  f^{\emph{OPE}}_{ij}(x) = F_{ij}^T\cdot P(x),
\end{equation}
where $P(x)$ is a vector composed of the values of the 2D-Fourier series at coordinate $x$.  Specifically,
$P(x)=[1, \sqrt{2}\cos(\pi x_1),\cdots,2\sin(k\pi x_1)\sin(k\pi x_2)]^T$. Here, $k$ is the selected maximum frequency.

In order to achieve a Rot-E OPE we need to first replace the encoder into a correlated Rot-E encoder. Then, we can proceed by setting
$\varphi$ in the input layer \eqref{input_layer} as
\begin{equation}\label{Exam_1}
    \varphi(W, F, x) = F^T\cdot P(x).
\end{equation}
Besides, we set $\psi$ in the output layer \eqref{output_layer} as an identity function, set all the elements in $W_{\outp}^1$ as $\frac{1}{t}$, and set the number of intermediate layers as $0$. In summary, the Rot-E OPE can be represented as
\begin{equation}\label{EQ_OPE}
  f_{ij}^{\emph{E-OPE}}(x)=\sum_{A\in S}\frac{1}{t} \(F^A_{ij}\)^T\cdot P\(A^{-1}x\),
\end{equation}
where $F$ here is provided by the Rot-E encoder.

\textbf{Rot-E LTE.} Local texture estimator (LTE) takes a dominant frequency estimator to construct INR \cite{chen2021learning}, which has achieved the SOTA performance in ASISR tasks. Formally, it can be calculated sd
\begin{equation}\label{LTE}
  f^{\emph{LTE}}_{ij}(x)  = \MLP_{M}\({\hat{F}_{ij}}\odot\begin{bmatrix}
                             \cos\(\pi \tilde{F}_{ij} x\) \\
                             \sin\(\pi \tilde{F}_{ij} x\)
                           \end{bmatrix} \),
\end{equation}
where $\hat{F}\in \mathbb{R}^{2K}$ and $\tilde{F}\in\mathbb{R}^{K\times 2}$ are two special latent codes produced by the encoder\footnote{$\hat{F}$ and  $\tilde{F}$ are usually achieved by adopting two convolutional network on the feature map  from the previous layer of the encoder, respectively.}, representing the amplitude vector and the frequency matrix, respectively.

In order to achieve a Rot-E LTE we should first replace the encoder into a correlated Rot-E encoder. Then,
let $F=\{\hat{F},  \tilde{F}\}$,
we only need to set the number of intermediate layers as $0$,  set
$\varphi$ in the input layer \eqref{input_layer} as
\begin{equation}\label{Exam_1}
    \varphi(W, F, x) = \hat{F}_{ij}\odot\begin{bmatrix}
                             \cos\(\pi \tilde{F}_{ij} x\) \\
                             \sin\(\pi \tilde{F}_{ij} x\)\end{bmatrix},
\end{equation}
and  set $\psi= \MLP_{M-1}$. In summary, the Rot-E LTE can be represented as
\begin{equation}\label{EQ_OPE}\begin{split}
&f_{ij}^{\emph{E-LTE}}(x)=\\
&\MLP_{M\!-\!1} \!\(\sum_{A,B\in S}\!\! W^1_{\outp}\! \cdot\! \hat{F}_{ij}^{B^{-1}\!A}\!\odot\!\begin{bmatrix}
                             \cos\(\pi \tilde{F}_{ij}^{B^{-1}\!A} x\) \\
                             \sin\(\pi \tilde{F}_{ij}^{B^{-1}\!A} x\)\end{bmatrix}\).
                              \end{split}
\end{equation}


\subsection{Rot-E Encoder Implementation and Improvement}
\subsubsection{Bicubic Parameterization for Rot-E Convolutions}

As shown in Fig. \ref{fig:method}, the Rot-E encoder is necessary for Rot-E ASISR. Fortunately, Rot-E convolutions applicable for low-level vision tasks have been proposed in recent years \cite{xie2022fourier,fu2023rotation}. These advances make it straightforward to transform traditional convolution-based encoders into Rot-E encoders.

The 2D filter parameterization technique has been shown to be crucial in designing Rot-E convolutions \cite{weiler2018learning,weiler2019general,xie2022fourier}, and a better representation accuracy will lead to better performance.
Generally, the following linear combination formulation is usually exploited for 2D filter parametrization,
\begin{equation}\label{general}
  {\phi}(x) = \sum_{k = 1}^K w_k \cdot\phi_{k}(x),
\end{equation}
where  ${\phi}$ is a to-be-learnt 2D latent function for a 2D filter,
$\{\phi_k, k = 1,2,\cdots, K\}$  denote the basis set with $K$ prespecified basis functions, and $w_k$ is the $k$-th representation weight.
Currently, the harmonic-basis-based filter parametrization \cite{weiler2018learning,weiler2019general} and the Fourier-basis-based  filter parametrization \cite{xie2022fourier}  have achieved SOTA performance in design Rot-E convolution of low-lever vision tasks, for their relatively high representation accuracy.

However, as shown in Figure \ref{fig:bicubic}, the continuous 2D functions achieved through harmonic-based and Fourier-expansion-based filter parametrization \cite{weiler2018learning,xie2022fourier} still exhibit a significant gap compared to the true underlying function of the 2D filters. This can be rationally explained by that these filter methods are highly dependent on high-frequency sine and cosine functions, which inevitably lead to many unnecessary function fluctuations.

In order to alleviate  this issue, we suggest a  bicubic-basis-based filtering  parametrization for Rot-E convolutions. Specifically,  the proposed filtering  parametrization inherits the formulation in \eqref{general}, while exploiting the functions in the following set as basis functions:
\begin{equation}\label{bicubic_basis}
  \l\{\!\Bic\(\frac{x_1}{\delta}\!-\!a\)\!\cdot\!\Bic\(\frac{x_2}{\delta}\!-\!b\)\mid a,\!b = -\!\frac{p\!-\!1}{2},\!\cdots\!, \frac{p\!-\!1}{2}\!\r\},
\end{equation}
where $x = [x_1, x_2]^T$ is the 2D coordinate,  $p$ is the filter size, $\delta$ is the mesh size, and $\Bic$ is the cubic interpolation function \cite{fadnavis2014image}, defined as
\begin{equation}\label{bicubic}
  \Bic(y) = \l\{ \!\!\begin{array}{ccc}
                  \!1.5|y|^3\!-\!2.5|y|^2\!+\!1 & \!\!\!\mbox{if}\!\!\! & |y|\!\leq\! 1 \\
                   \!\!-0.5|y|^3\!+\!2.5|y|^2\!-\!4|y|\!+\!2 & \!\!\!\mbox{if}\!\!\! & 1\!<\!|y|\!\leq\! 2 \\
                   \!0 & \!\!\!\mbox{if}\!\!\! & 2\!<\!|y|
                 \end{array}\!\!.\!\!\!\!\r.
\end{equation}

It is easy to find that representing the latent function of a filter with this bicubic-basis-based filter parametrization is  equivalent to obtaining the implicit function of the filter kernel through bicubic interpolation of the elements of the filter kernel.
This is more natural than previous methods based on the harmonic \cite{weiler2018learning} or Fourier expansion \cite{xie2022fourier},  since bilinear interpolation is the most commonly used interpolation method in image transformation, and has been effectively applied in practical scenarios.  It can be observed in Fig. \ref{fig:bicubic}(e) that the latent function represented with bicubic-basis-based filter parametrization is closer to the true latent 2D continuous filter.

In this paper, we incorporate this bicubic-basis-based filter parametrization  into the framework of previous Rot-E convolutions \cite{weiler2018learning,weiler2019general,xie2022fourier}, replacing the original filter parametrization method, resulting in a new Rot-E convolutional method. We name it B-Conv, and exploit it to construct Rot-E ASISR.

\subsubsection{Equivariant Settings for Transformer-Based Encoder}

The transformer-based encoder has been shown to be effective for ASISR tasks \cite{lee2022local, liang2021swinir, dosovitskiy2020image}. Therefore,  it is necessary to adopt Rot-E settings for transformers when constructing Rot-E ASISR methods.

\red{However, current Rot-E  Vision-Transformers \cite{he2021efficient, romero2020group, xu20232, hutchinson2021lietransformer} are mainly designed for classification tasks, while not suitable to low-level vision. Particularly, it is difficult to adapt  them to  shift-window version ones, i.e., Swin-Transformers \cite{liang2021swinir}, which  makes them almost  inapplicable for ASISR tasks. Besides, the complex designs of these methods can also interfere with the fair performance comparison between the proposed Rot-E INR and traditional INRs. Therefore, in this study, we construct a new Rot-E transformer by replacing the non-equivariant modules in the Vision Transformer \cite{liang2021swinir} with their Rot-E counterparts. }

Specifically, we replace the patches embedding layer with the Rot-E convolution layer,  replace all the linear layer with the Rot-E linear layer (i.e., the Rot-E convolution layer with kernel size of $1\times 1$), remove the position embedding module (which will break the intrinsic equivariance property of the entire network), and avoid to introduce any additional module. It can be proved that 

\begin{Rem}\label{Remark}
When $t = 4$, the simplified Vision-Transformer-based encoder and Swin-Transformer-based encoder constructed through the above manner are 
exactly rotation eqivariant on p4 rotation group (i.e., on $\frac{k\pi}{2}, k=1,\cdots,4$ rotations).
\end{Rem}


Remark 1 demonstrates that the transformer-based encoder, configured with our settings, exhibits precise rotation equivariance at typical discrete orientation degrees. This achieves a theoretical level comparable to established research \cite{shen2020pdo}. Proving its Rot-E property across arbitrary orientations remains a challenging task and worth investing in future research.


\begin{figure*}
    \centering
    \includegraphics[width=\linewidth]{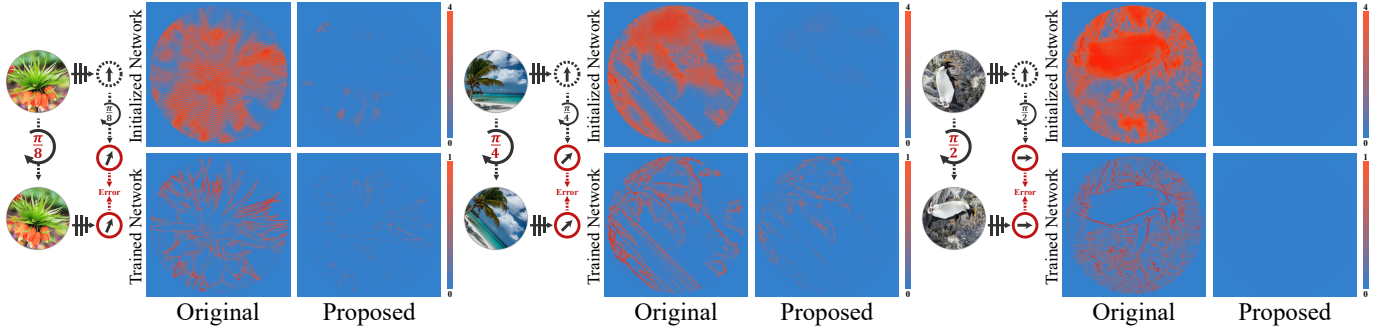}\vspace{-1.5mm}
    \caption{Illustration of rotation equivariant errors ($|\Phi[ \pi_{\tilde{A}}^I](I) - \pi_{\tilde{A}}^F[\Phi](I)|$ for image $I$ and network $\Phi$) of original LTE and its rotation-equivariant improvement (p16 rotation equivariant). The rotation degree is $\frac{\pi}{8}$ (left), $\frac{\pi}{4}$ (middle), and $\frac{\pi}{2}$ (right), respectively. The upper and bottom lines are the equivariant error map of randomly initialized networks without training and network trained for 100 epoches, respectively.}
    \label{fig:EQerror}\vspace{-0mm}
\end{figure*}

\begin{table*}[t]
  \caption{Average equivariant error of SR models on p4 rotation group, the results are averaged on 100 images in the DIV2K testing set.} \vspace{-1.5mm}
  \label{EQmeasure}
  \centering \setlength{\tabcolsep}{5.7pt}
  \begin{tabular}{lcccccccc}
    \toprule
                    &   \multicolumn{4}{c}{ $\times$4}          &  \multicolumn{4}{c}{ $\times$8}\\
                    \cmidrule(r){2-5} \cmidrule(r){6-9} 
     {Method}       &   \multicolumn{2}{c}{ Randomly initialization }          &  \multicolumn{2}{c}{ Trained network } &   \multicolumn{2}{c}{ Randomly initialization }          &  \multicolumn{2}{c}{ Trained network } \\
                    \cmidrule(r){2-3}\cmidrule(r){4-5} \cmidrule(r){6-7}\cmidrule(r){8-9} 
                     &   NMSE $\downarrow$&      NMAE$\downarrow$    &   {NMSE $\downarrow$}&      { NMAE $\downarrow$}     &   NMSE $\downarrow$&       NMAE $\downarrow$    &   {NMSE $\downarrow$}&      { NMAE $\downarrow$}\\
 \midrule  
   LIIF \cite{chen2021learning} & 1.025 $\!\pm\!$ 0.154 & 0.981 $\!\pm\!$ 0.087 & 0.058 $\!\pm\!$ 0.025 & 0.197 $\!\pm\!$ 0.044 & 1.021 $\!\pm\!$ 0.159 & 0.980 $\!\pm\!$ 0.089 & 0.060 $\!\pm\!$ 0.026 & 0.200 $\!\pm\!$ 0.044  \\ 
                        LIIF-EQ & \textbf{2e-03 $\!\pm\!$ 8e-04} & \textbf{4e-02 $\!\pm\!$ 9e-03 }& \textbf{2e-04 $\!\pm\!$ 6e-05} & \textbf{1e-02 $\!\pm\!$ 2e-03 }& \textbf{2e-03 $\!\pm\!$ 9e-04} & \textbf{4e-02 $\!\pm\!$ 1e-02 }& \textbf{2e-04 $\!\pm\!$ 6e-05} & \textbf{1e-02 $\!\pm\!$ 2e-03 } \\ 
 \midrule  
        OPE \cite{song2023ope}  & 1.445 $\!\pm\!$ 0.034 & 1.125 $\!\pm\!$ 0.023 & 0.055 $\!\pm\!$ 0.024 & 0.198 $\!\pm\!$ 0.045 & 1.453 $\!\pm\!$ 0.033 & 1.128 $\!\pm\!$ 0.022 & 0.056 $\!\pm\!$ 0.024 & 0.199 $\!\pm\!$ 0.045  \\ 
                         OPE-EQ & \textbf{4e-04 $\!\pm\!$ 8e-05} & \textbf{2e-02 $\!\pm\!$ 2e-03 }& \textbf{1e-04 $\!\pm\!$ 5e-05} & \textbf{8e-03 $\!\pm\!$ 1e-03 }& \textbf{4e-04 $\!\pm\!$ 7e-05} & \textbf{2e-02 $\!\pm\!$ 2e-03 }& \textbf{1e-04 $\!\pm\!$ 5e-05} & \textbf{8e-03 $\!\pm\!$ 1e-03 } \\ 
 \midrule  
        LTE \cite{lee2022local} & 1.274 $\!\pm\!$ 0.121 & 1.103 $\!\pm\!$ 0.064 & 0.352 $\!\pm\!$ 0.101 & 0.428 $\!\pm\!$ 0.084 & 1.249 $\!\pm\!$ 0.111 & 1.092 $\!\pm\!$ 0.060 & 0.316 $\!\pm\!$ 0.091 & 0.407 $\!\pm\!$ 0.079  \\ 
                         LTE-EQ & \textbf{1e-03 $\!\pm\!$ 5e-04} & \textbf{4e-02 $\!\pm\!$ 6e-03 }& \textbf{1e-04 $\!\pm\!$ 5e-05} & \textbf{1e-02 $\!\pm\!$ 2e-03 }& \textbf{1e-03 $\!\pm\!$ 4e-04} & \textbf{3e-02 $\!\pm\!$ 6e-03 }& \textbf{1e-04 $\!\pm\!$ 5e-05} & \textbf{1e-02 $\!\pm\!$ 2e-03 } \\ 
 \bottomrule  
\end{tabular}\vspace{-0mm}
\end{table*}

\begin{table*}[t]
  \caption{Average equivariant error of the randomly initialized models on p8 and p16 rotation group,  averaging on 100 images in the DIV2K testing set.} \vspace{-1.5mm}
  \label{EQmeasure2}
  \centering \setlength{\tabcolsep}{6.3pt}
  \begin{tabular}{lcccccccc}
    \toprule
                    &   \multicolumn{4}{c}{ $\times$4}          &  \multicolumn{4}{c}{ $\times$8}\\
                    \cmidrule(r){2-5} \cmidrule(r){6-9} 
     {Method}     &   \multicolumn{2}{c}{Random $\nicefrac{k\pi}{4}$ rotation}          &  \multicolumn{2}{c}{Random $\nicefrac{k\pi}{8}$ rotation} &   \multicolumn{2}{c}{Random $\nicefrac{k\pi}{4}$ rotation}          &  \multicolumn{2}{c}{Random $\nicefrac{k\pi}{8}$ rotation} \\
                    \cmidrule(r){2-3}\cmidrule(r){4-5} \cmidrule(r){6-7}\cmidrule(r){8-9} 
                     &   NMSE $\downarrow$&      NMAE$\downarrow$    &   {NMSE $\downarrow$}&      { NMAE $\downarrow$}     &   NMSE $\downarrow$&       NMAE $\downarrow$    &   {NMSE $\downarrow$}&      { NMAE $\downarrow$}\\
 \midrule  
   LIIF \cite{chen2021learning} & 1.015 $\!\pm\!$ 0.163 & 0.944 $\!\pm\!$ 0.095 & 0.982 $\!\pm\!$ 0.209 & 0.927 $\!\pm\!$ 0.115 & 1.013 $\!\pm\!$ 0.165 & 0.942 $\!\pm\!$ 0.097 & 0.981 $\!\pm\!$ 0.209 & 0.925 $\!\pm\!$ 0.115  \\ 
                        LIIF-EQ & \textbf{0.058 $\!\pm\!$ 0.045} & \textbf{0.186 $\!\pm\!$ 0.111} & \textbf{0.063 $\!\pm\!$ 0.034} & \textbf{0.210 $\!\pm\!$ 0.081} & \textbf{0.071 $\!\pm\!$ 0.054} & \textbf{0.207 $\!\pm\!$ 0.126} & \textbf{0.078 $\!\pm\!$ 0.040} & \textbf{0.236 $\!\pm\!$ 0.090}  \\ 
 \midrule  
        OPE \cite{song2023ope}  & 1.349 $\!\pm\!$ 0.077 & 1.123 $\!\pm\!$ 0.058 & 1.369 $\!\pm\!$ 0.064 & 1.139 $\!\pm\!$ 0.048 & 1.355 $\!\pm\!$ 0.062 & 1.114 $\!\pm\!$ 0.043 & 1.371 $\!\pm\!$ 0.051 & 1.124 $\!\pm\!$ 0.036  \\ 
                         OPE-EQ & \textbf{0.745 $\!\pm\!$ 0.497} & \textbf{0.704 $\!\pm\!$ 0.456} & \textbf{0.934 $\!\pm\!$ 0.359} & \textbf{0.879 $\!\pm\!$ 0.326} & \textbf{0.815 $\!\pm\!$ 0.540} & \textbf{0.739 $\!\pm\!$ 0.478} & \textbf{1.024 $\!\pm\!$ 0.386} & \textbf{0.924 $\!\pm\!$ 0.340}  \\ 
 \midrule  
        LTE \cite{lee2022local} & 1.096 $\!\pm\!$ 0.102 & 1.016 $\!\pm\!$ 0.061 & 1.089 $\!\pm\!$ 0.110 & 1.011 $\!\pm\!$ 0.065 & 1.059 $\!\pm\!$ 0.116 & 0.991 $\!\pm\!$ 0.066 & 1.049 $\!\pm\!$ 0.119 & 0.986 $\!\pm\!$ 0.068  \\ 
                         LTE-EQ & \textbf{0.107 $\!\pm\!$ 0.081} & \textbf{0.253 $\!\pm\!$ 0.158} & \textbf{0.127 $\!\pm\!$ 0.064} & \textbf{0.303 $\!\pm\!$ 0.115} & \textbf{0.104 $\!\pm\!$ 0.079} & \textbf{0.247 $\!\pm\!$ 0.156} & \textbf{0.128 $\!\pm\!$ 0.064} & \textbf{0.300 $\!\pm\!$ 0.116}  \\ 
 \bottomrule   
\end{tabular}\vspace{-0mm}
\end{table*}

\begin{figure*}
    \centering
    \includegraphics[width=0.81\linewidth]{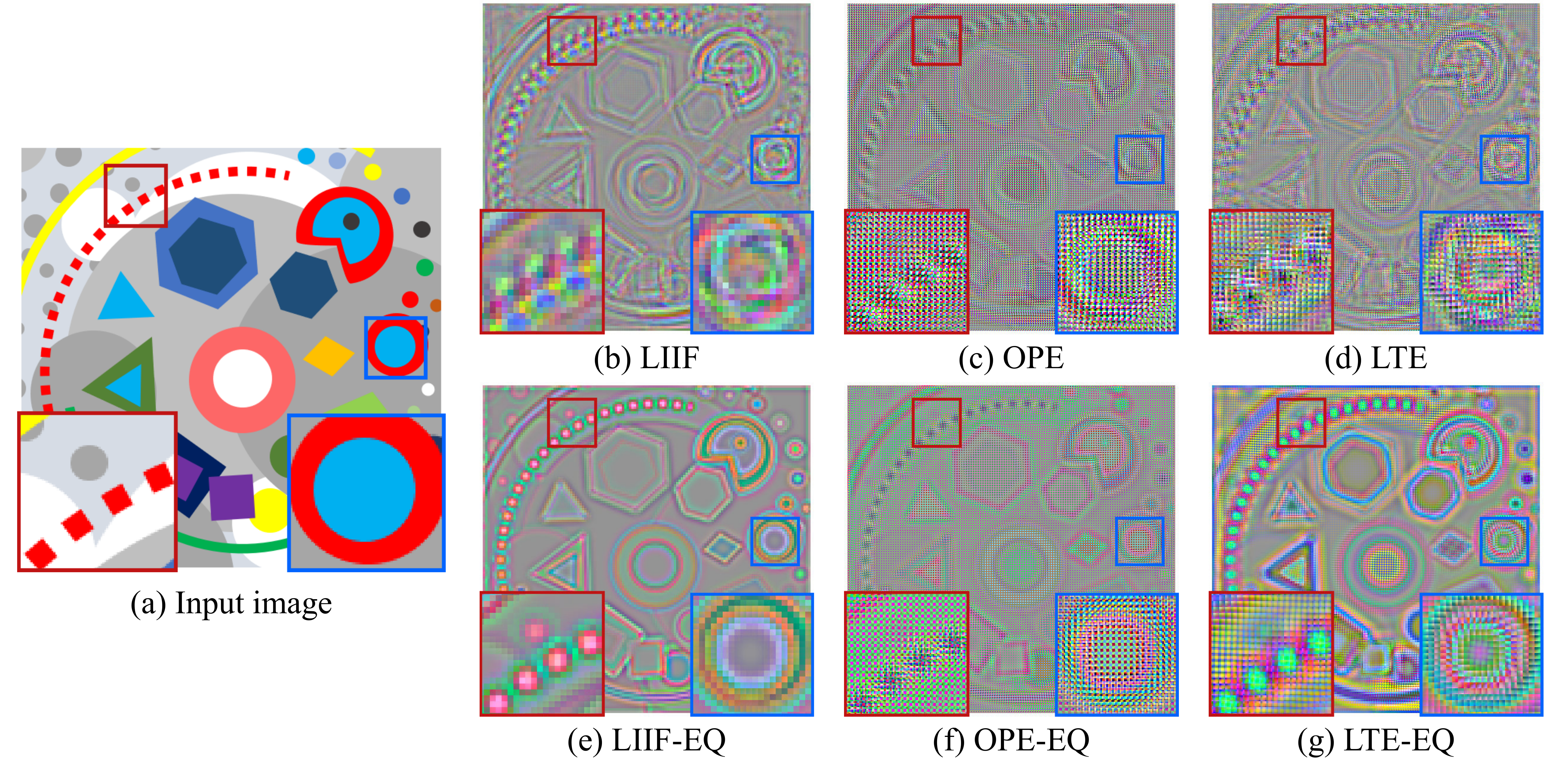}\vspace{-2mm}
    \caption{\red{Illustration of the output images of original  and its rotation-equivariant improvement (p16 rotation equivariant), when the network is randomly initialized without any training. (a) An input image  composed of typical geometric shapes. (b)-(g) Output images of the original method and our rotation-equivariant-improved method, resprectively.} }
    \label{fig:EQexam}\vspace{-2mm}
\end{figure*}

\section{Experimental results}

In this section, we first conduct simulated experiments to empirically evaluate the equivariance error of the proposed method. Then, we present experimental results that demonstrate the superiority of the proposed framework when being combined with typical ASISR methods in a plug-and-play manner. Finally, we performed an ablation study to verify the effectiveness of each proposed module.

\subsection{Equivariance Verification}

In Fig. 2, we have shown the capability of the proposed framework in keeping rotation symmetry. 
We also have provided theoretical analysis of the equivariance error associated with the proposed module in Sec. 3.2. Here, we further empirically investigate the equivariance error of the proposed module. 

We first compare the equivariance error of the original networks and their rotation equivariant versions constructed with the proposed framework. We exploited three typical INRs for the experiments, including LIIF \cite{chen2021learning}, OPE \cite{song2023ope}, and LTE \cite{lee2022local}. We chose the ERSD network as encoder, with 16 layers and 64 channels for each layer. 
All the competing methods are trained on with DIV2K dataset \cite{timofte2017ntire},   and we
exploit 100  images from the testing set of the DIV2K dataset \cite{timofte2017ntire} to test the equivariance error. 
We exploit two quality indices for performance evaluation. The first is the most typical quality indice, i.e.,   normalized mean square error (NMSE) and  the second is normalized mean absolute error (NMAE)\footnote{NMSE and NMAE are respectively calculated by $\nicefrac{\left\|x_r-x_0\right\|_2}{\left\|x_0\right\|_2}$ and  $\nicefrac{\left\|x_r-x_0\right\|_1}{\left\|x_0\right\|_1}$,  where $x_0$ the rotated output of original image and $x_r$ is the output of the rotated image.}.  It should be noted that NMAE is consistent with the metric used in the theoretical analysis.

Table \ref{EQmeasure} shows the average equivariant error of different SR models on p4 rotation group ($\frac{k\pi}{2} \mbox{ rotaions, } k = 1,2,\cdots, 4$). We can find that the equivariant error of the methods improved with the proposed Rot-E framework is much closer to zero compared to those of the original ones. Similar results can also be observed in the right part of Fig. \ref{fig:EQerror}, when the rotation degree is $\frac{\pi}{2}$, the equivariant error of the proposed method is closer to zero.
These results are consistent with the theoretical results in Corollary 3 that equivalence is strict when $t=4$, the small error arises from a small constant (about $10^{-7}$) added to the coordinates in the local ensemble process \cite{chen2021learning}.
Table \ref{EQmeasure2} further shows the average equivariant error of the randomly initialized models on p8 ($\frac{k\pi}{4} \mbox{ rotaions, } k = 1,2,\cdots, 8$) and p16 ($\frac{k\pi}{8} \mbox{ rotaions, } k = 1,2,\cdots, 16$) rotation group.  We can observe that the equivariant error of the methods improved with the proposed Rot-E framework is also much lower than the original ones in these cases.  Similar results can also be observed in the left and middle cases in Fig. \ref{fig:EQerror}.

\red{Fig. \ref{fig:EQexam} shows the output images of the original methods and the proposed Rot-E improvements (p16 rotation equivariant), when the networks were randomly initialized without undergoing any training. We chose an input image  composed of typical geometric shapes for easy observing the rotation equivariance.  It can be easily seen that the outputs of different INRs are indeed very different from each other, while the proposed method can consistently help making them more rotation equivariant. Specifically, the outputs of original methods seem too chaotic to finely deliver the rotational symmetry of the geometric shapes. 
In comparison, the results of the proposed ameliorations can better maintain rotational symmetry with more orderly structural patterns. It can be observed that when the input local patten is rotated, the output local latent function will rotate accordingly in the proposed method. These results intuitively verify the rotation equivariance characteristics of the proposed framework.}


\begin{table*}[ht]
  \centering 
    \caption{Average PSNR (dB) of the \textbf{in-scale super-resolution} results obtained by all comparison methods on different benchmark datasets.}
    \centering \setlength{\tabcolsep}{5pt}
\begin{tabular}{c c c c c c c c c c c c c c c c}
    \toprule
     \multirow{3}{*}{Endoder} & \multirow{3}{*}{INR} & \multirow{3}{*}{Mode} & \multicolumn{3}{c}{Urban100 \cite{huang2015single}} & \multicolumn{3}{c}{BSD100 \cite{martin2001database}} & \multicolumn{3}{c}{Set14 \cite{zeyde2010single}} & \multicolumn{3}{c}{Set5 \cite{bevilacqua2012low}} & \multirow{2}{*}{Param. }\\
     \cmidrule(r){4-6}\cmidrule(r){7-9}\cmidrule(r){10-12}\cmidrule(r){13-15}
     && & x2 & x3 & x4 & x2 & x3 & x4 & x2 & x3 & x4 & x2 & x3 & x4 \\
     \midrule
     \multirow{6}{*}{$\begin{array}{c}
                        \mbox{EDSR \cite{lim2017enhanced}} \\
                        \mbox{(baseline)} 
                      \end{array}$} 
     & \multirow{2}{*}{LIIF \cite{chen2021learning}} 
      &Original& 32.18 & 28.24 & 26.18 & 32.18 & 29.12 & 27.61 & 33.63 & 30.34 & 28.64 & 38.00 & 34.41 & 32.22 & 1.6M \\
     &&Proposed& \textbf{32.39} & \textbf{28.36} & \textbf{26.28} & \textbf{32.21} & \textbf{29.14} & \textbf{27.62} & \textbf{33.68} & \textbf{30.28} & \textbf{28.67} & \textbf{38.01} & \textbf{34.44} & \textbf{32.29} & {1.3M} \\
     \cmidrule(r){2-3}\cmidrule(r){4-6}\cmidrule(r){7-9}\cmidrule(r){10-12}\cmidrule(r){13-15}\cmidrule(r){16-16}
     & \multirow{2}{*}{OPE \cite{song2023ope}} 
      &Original& 31.93 & 28.06 & 25.93 & 32.09 & 29.03 & 27.52 & 33.54 & 30.24 & 28.53 & 37.90 & 34.17 & 31.98 & 1.3M \\
     &&Proposed& \textbf{32.00} & \textbf{28.15} & \textbf{25.98} & \textbf{32.14} & \textbf{29.06} & \textbf{27.55} & \textbf{33.55} & \textbf{30.27} & \textbf{28.56} & \textbf{37.95} & \textbf{34.28} & \textbf{32.06} & {1.1M}\\
     \cmidrule(r){2-3}\cmidrule(r){4-6}\cmidrule(r){7-9}\cmidrule(r){10-12}\cmidrule(r){13-15}\cmidrule(r){16-16}
     & \multirow{2}{*}{LTE \cite{lee2022local}} 
      &Original& 32.28 & 28.32 & 26.23 & 32.21 & 29.14 & 27.62 & 33.68 & 30.38 & 28.66 & 38.03 & 34.50 & 32.27 & 1.7M \\
     &&Proposed& \textbf{32.43} & \textbf{28.37} & \textbf{26.31} & \textbf{32.23} & \textbf{29.17} & \textbf{27.64} & \textbf{33.74} & \textbf{30.42} & \textbf{28.69} & \textbf{38.07} & \textbf{34.51} & \textbf{32.28} & {1.2M} \\
          \midrule
     \multirow{6}{*}{RDN \cite{zhang2018residual}}
     &  \multirow{2}{*}{LIIF \cite{chen2021learning}} 
      &Original& 32.90 & 28.86 & 26.70 & 32.33 & 29.27 & 27.75 & 33.97 & 30.54 & 28.80 & 38.19 & 34.69 & 32.50 & 22.3M \\
     &&Proposed& \textbf{33.06} & \textbf{28.99} & \textbf{26.81} & \textbf{32.34} & \textbf{29.30} & \textbf{27.78} & \textbf{34.05} & \textbf{30.60} & \textbf{28.88} & \textbf{38.20} & \textbf{34.77} & \textbf{32.61} & {15.4M} \\
     \cmidrule(r){2-3}\cmidrule(r){4-6}\cmidrule(r){7-9}\cmidrule(r){10-12}\cmidrule(r){13-15}\cmidrule(r){16-16}
     & \multirow{2}{*}{OPE \cite{song2023ope}} 
      &Original& 32.65 & 28.74 & 26.56 & 32.24 & 29.17 & 27.68 & 33.85 & 30.49 & 28.76 & 38.12 & 34.47 & 32.27 & 22.1M \\
     &&Proposed& \textbf{32.75} & \textbf{28.81} & \textbf{26.61} & \textbf{32.28} & \textbf{29.21} & \textbf{27.72} & \textbf{33.90} & \textbf{30.52} & \textbf{28.80} & \textbf{38.13} & \textbf{34.57} & \textbf{32.32} & {15.2M} \\
     \cmidrule(r){2-3}\cmidrule(r){4-6}\cmidrule(r){7-9}\cmidrule(r){10-12}\cmidrule(r){13-15}\cmidrule(r){16-16}
     & \multirow{2}{*}{LTE \cite{lee2022local}} 
      &Original& 33.03 & 28.96 & 26.81 & 32.36 & 29.30 & 27.78 & 34.04 & 30.60 & 28.88 & 38.24 & 34.77 & 32.56 & 22.5M \\
     &&Proposed& \textbf{33.23} & \textbf{29.05} & \textbf{26.91} & \textbf{32.40} & \textbf{29.33} & \textbf{27.79} & \textbf{34.21} & \textbf{30.63} & \textbf{28.92} & \textbf{38.28} & \textbf{34.82} & \textbf{32.68} & {15.3M} \\
     \midrule
     \multirow{6}{*}{SwinIR \cite{liang2021swinir}}
     &  \multirow{2}{*}{LIIF \cite{chen2021learning}} 
      &Original& 33.34 & 29.31 & 27.12 & 32.40 & 29.35 & 27.83 & 34.10 & 30.73 & 28.99 & 38.28 & 34.84 & 32.71 & 13.5M \\
     &&Proposed& \textbf{33.54} & \textbf{29.44} & \textbf{27.22} & \textbf{32.44} & \textbf{29.39} & \textbf{27.87} & \textbf{34.36} & \textbf{30.78} & \textbf{29.02} & \textbf{38.34} & \textbf{34.89} & \textbf{32.74} & {7.8M} \\
     \cmidrule(r){2-3}\cmidrule(r){4-6}\cmidrule(r){7-9}\cmidrule(r){10-12}\cmidrule(r){13-15}\cmidrule(r){16-16}
     & \multirow{2}{*}{OPE \cite{song2023ope}} 
      &Original& 33.12 & 29.12 & 26.89 & 32.35 & 29.26 & 27.75 & 34.10 & 30.66 & 28.90 & 38.25 & 34.68 & 32.49 & 13.3M \\
     &&Proposed& \textbf{33.35} & \textbf{29.34} & \textbf{27.09} & \textbf{32.41} & \textbf{29.32} & \textbf{27.81} & \textbf{34.12} & \textbf{30.73} & \textbf{28.97} & \textbf{38.32} & \textbf{34.76} & \textbf{32.55} & {7.7M} \\
     \cmidrule(r){2-3}\cmidrule(r){4-6}\cmidrule(r){7-9}\cmidrule(r){10-12}\cmidrule(r){13-15}\cmidrule(r){16-16}
     & \multirow{2}{*}{LTE \cite{lee2022local}} 
      &Original& 33.46 & 29.38 & 27.23 & 32.45 & 29.38 & 27.85 & 34.21 & 30.78 & 29.02 & 38.32 & 34.89 & 32.76 & 13.7M \\
     &&Proposed& \textbf{33.67} & \textbf{29.51} & \textbf{27.33} & \textbf{32.47} & \textbf{29.42} & \textbf{27.90} & \textbf{34.33} & \textbf{30.83} & \textbf{29.04} & \textbf{38.38} & \textbf{34.94} & \textbf{32.78} & {7.7M} \\
     
    \bottomrule
  \end{tabular}
  \label{tab:in-scale}
  \vspace{0mm}
\end{table*}

\begin{figure*}
	\centering
	\includegraphics[width=0.98\linewidth]{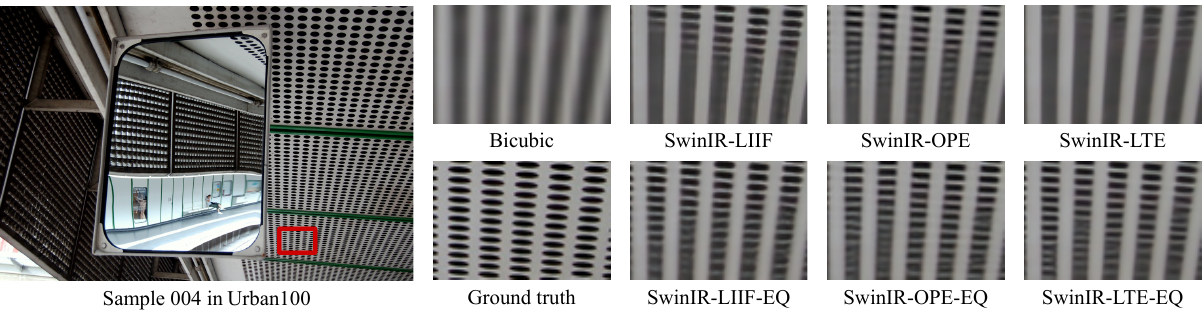}\vspace{-2mm}
	\caption{Visual comparison of the \textbf{in-scale super resolution} ($\times 4$) results  from various methods on \textit{img004} of the Urban100 \cite{huang2015single} datasets.}
	\label{fig:exp_in_scale} \vspace{-1mm}
\end{figure*}

\begin{table*}[ht]
  \centering 
    \caption{Average PSNR (dB) of the \textbf{out-scale super-resolution}  results obtained by all comparison methods on different benchmark datasets.}
    \centering \setlength{\tabcolsep}{5pt}
\begin{tabular}{c c c c c c c c c c c c c c c c}
    \toprule
     \multirow{3}{*}{Encoder} & \multirow{3}{*}{INR} & \multirow{3}{*}{Mode} & \multicolumn{3}{c}{Urban100 \cite{huang2015single}} & \multicolumn{3}{c}{BSD100 \cite{martin2001database}} & \multicolumn{3}{c}{Set14 \cite{zeyde2010single}} & \multicolumn{3}{c}{Set5 \cite{bevilacqua2012low}} & \multirow{2}{*}{Param. }\\
     \cmidrule(r){4-6}\cmidrule(r){7-9}\cmidrule(r){10-12}\cmidrule(r){13-15}
     && & x6 & x8 & x12 & x6 & x8 & x12 & x6 & x8 & x12 & x6 & x8 & x12 \\
     \midrule
     \multirow{6}{*}{$\begin{array}{c}
                        \mbox{EDSR \cite{lim2017enhanced}} \\
                        \mbox{(baseline)} 
                      \end{array}$} 
     & \multirow{2}{*}{LIIF \cite{chen2021learning}} 
      &Original& 23.81 & 22.48 & 20.92 & 25.84 & 24.79 & 23.54 & 26.48 & 24.96 & 23.17 & 28.89 & \textbf{26.98} & \textbf{24.51} & 1.6M \\
     &&Proposed& \textbf{23.89} & \textbf{22.52} & \textbf{20.95} & \textbf{25.87} & \textbf{24.81} & \textbf{23.55} & \textbf{26.50} & \textbf{24.99} & \textbf{23.21} & \textbf{28.90} & {26.95} & {24.50} & {1.3M} \\
     \cmidrule(r){2-3}\cmidrule(r){4-6}\cmidrule(r){7-9}\cmidrule(r){10-12}\cmidrule(r){13-15}\cmidrule(r){16-16}
     & \multirow{2}{*}{OPE \cite{song2023ope}} 
      &Original& 23.63 & 22.36 & 20.86 & 25.78 & 24.74 & 23.50 & 26.31 & 24.82 & 23.17 & 28.65 & 26.78 & 24.43 & 1.3M \\
     &&Proposed& \textbf{23.68} & \textbf{22.41} & \textbf{20.90} & \textbf{25.79} & \textbf{24.75} & \textbf{23.51} & \textbf{26.35} & \textbf{24.86} & \textbf{23.18} & \textbf{28.77} & \textbf{26.78} & \textbf{24.47} & {1.1M} \\
     \cmidrule(r){2-3}\cmidrule(r){4-6}\cmidrule(r){7-9}\cmidrule(r){10-12}\cmidrule(r){13-15}\cmidrule(r){16-16}
     & \multirow{2}{*}{LTE \cite{lee2022local}} 
      &Original& 23.85 & 22.50 & 20.91 & 25.86 & 24.81 & 23.52 & 26.47 & 24.95 & \textbf{23.18} & \textbf{28.98} & 26.97 & \textbf{24.46} & 1.7M \\
     &&Proposed& \textbf{23.92} & \textbf{22.55} & \textbf{20.94} & \textbf{25.87} & \textbf{24.81} & \textbf{23.53} & \textbf{26.50} & \textbf{24.96} & {23.17} & \textbf{28.98} & \textbf{26.98} & {24.43} & {1.2M} \\ 
     \midrule
     \multirow{6}{*}{RDN \cite{zhang2018residual}}
     &  \multirow{2}{*}{LIIF \cite{chen2021learning}} 
      &Original& 24.21 & 22.79 & 21.15 & 25.99 & 24.93 & 23.63 & 26.65 & 25.14 & \textbf{23.32} & 29.21 & 27.17 & \textbf{24.75} & 22.3M \\
     &&Proposed& \textbf{24.29} & \textbf{22.89} & \textbf{21.24} & \textbf{26.01} & \textbf{24.93} & \textbf{23.64} & \textbf{26.70} & \textbf{25.19} & {23.29} & \textbf{29.34} & \textbf{27.29} & {24.70} & {15.4M} \\
     \cmidrule(r){2-3}\cmidrule(r){4-6}\cmidrule(r){7-9}\cmidrule(r){10-12}\cmidrule(r){13-15}\cmidrule(r){16-16}
     & \multirow{2}{*}{OPE \cite{song2023ope}} 
      &Original& 24.10 & 22.75 & 21.15 & 25.94 & 24.91 & 23.62 & 26.56 & 25.06 & 23.31 & 28.95 & 27.00 & 24.62 & 22.1M \\
     &&Proposed& \textbf{24.16} & \textbf{22.80} & \textbf{21.20} & \textbf{25.97} & \textbf{24.92} & \textbf{23.63} & \textbf{26.61} & \textbf{25.10} & \textbf{23.34} & \textbf{29.04} & \textbf{27.10} & \textbf{24.65} & {15.2M} \\
     \cmidrule(r){2-3}\cmidrule(r){4-6}\cmidrule(r){7-9}\cmidrule(r){10-12}\cmidrule(r){13-15}\cmidrule(r){16-16}
     & \multirow{2}{*}{LTE \cite{lee2022local}} 
      &Original& 24.31 & 22.84 & 21.10 & 26.01 & 24.94 & \textbf{23.60} & 26.69 & \textbf{25.16} & \textbf{23.32} & 29.22 & \textbf{27.18} & \textbf{24.66} & 22.5M \\
     &&Proposed& \textbf{24.39} & \textbf{22.90} & \textbf{21.15} & \textbf{26.03} & \textbf{24.93} & {23.58} & \textbf{26.71} & \textbf{25.16} & {23.27} & \textbf{29.28} & \textbf{27.18} & {24.56} & {15.3M} \\
     \midrule
     \multirow{6}{*}{SwinIR \cite{liang2021swinir}}
     &  \multirow{2}{*}{LIIF \cite{chen2021learning}} 
      &Original& 24.60 & 23.14 & 21.45 & 26.07 & 25.01 & 23.63 & 26.78 & 25.37 & 23.36 & 29.46 & 27.38 & 24.96 & 13.5M \\
     &&Proposed& \textbf{24.66} & \textbf{23.20} & \textbf{21.48} & \textbf{26.10} & \textbf{25.03} & \textbf{23.65} & \textbf{26.90} & \textbf{25.39} & \textbf{23.45} & \textbf{29.52} & \textbf{27.41} & \textbf{24.99} & {7.8M} \\
     \cmidrule(r){2-3}\cmidrule(r){4-6}\cmidrule(r){7-9}\cmidrule(r){10-12}\cmidrule(r){13-15}\cmidrule(r){16-16}
     & \multirow{2}{*}{OPE \cite{song2023ope}} 
      &Original& 24.37 & 23.00 & 21.37 & 26.02 & 24.96 & 23.61 & 26.69 & 25.26 & 23.37 & 29.34 & 27.26 & 24.88 & 13.3M \\
     &&Proposed& \textbf{24.54} & \textbf{23.13} & \textbf{21.46} & \textbf{26.06} & \textbf{25.01} & \textbf{23.66} & \textbf{26.79} & \textbf{25.30} & \textbf{23.41} & \textbf{29.39} & \textbf{27.36} & \textbf{24.93} & {7.7M} \\
     \cmidrule(r){2-3}\cmidrule(r){4-6}\cmidrule(r){7-9}\cmidrule(r){10-12}\cmidrule(r){13-15}\cmidrule(r){16-16}
     & \multirow{2}{*}{LTE \cite{lee2022local}} 
      &Original& 24.66 & 23.15 & 21.42 & 26.08 & 25.02 & \textbf{23.63} & 26.84 & 25.37 & \textbf{23.39} & 29.51 & 27.37 & \textbf{25.05} & 13.7M \\
     &&Proposed& \textbf{24.71} & \textbf{23.18} & \textbf{21.43} & \textbf{26.11} & \textbf{25.02} & {23.62} & \textbf{26.87} & \textbf{25.37} & {23.38} & \textbf{29.56} & \textbf{27.43} & {24.94} & {7.7M} \\
    \bottomrule
  \end{tabular}
  \label{tab:out-scale}
  \vspace{-0mm}
\end{table*}

\begin{figure*}
	\centering
	\includegraphics[width=0.98\linewidth]{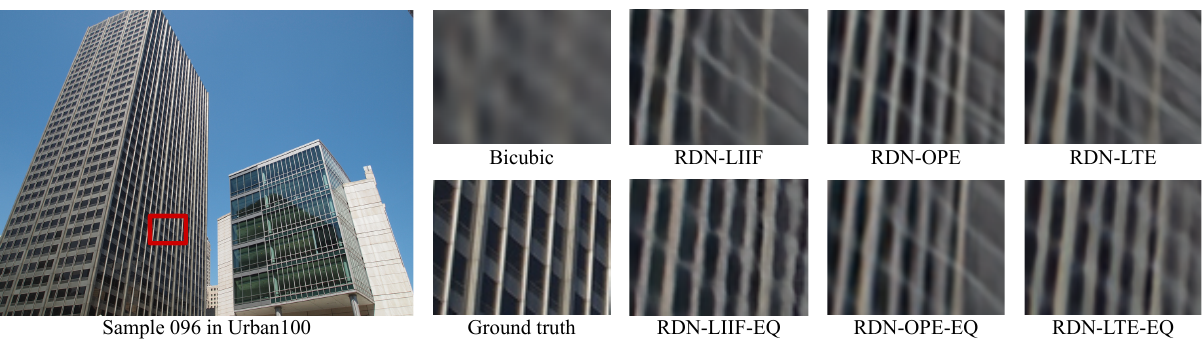} \vspace{-2mm}
	\caption{Visual comparison of the \textbf{out-scale super resolution}  ($\times 8$) results from various methods on \textit{img096} of the Urban100 \cite{huang2015single} datasets.}
	\label{fig:exp_out_scale}\vspace{-0mm}
\end{figure*}


\subsection{Plug and Play with typical ASISR methods}

We then test the propose method on 3 typical encoders and 3 typical INR, respectively. Specifically,  we utilize the encoder of EDSR (basedline version, with 16 residual blocks and 64 channels for each feature map) \cite{lim2017enhanced}, RDN \cite{zhang2018residual} and SwinIR \cite{liang2021swinir}, representing the light-weight, SOTA-CNN-based and transformer-based encoders, respectively. For INR, the proposed method is test on LIIF \cite{chen2021learning}, OPE \cite{song2023ope}, LTE \cite{lee2022local}, representing classical MLP-based INR, parameter-free and SOTA INRs for ASISR, respectively. For all combinations of these encoders and INRs, we compared the performance of the original non-equivariant versions with those of the Rot-E versions implemented based on the proposed method.

{\bf Datasets.}  Following previous works \cite{chen2021learning, lee2022local, cao2023ciaosr, song2023ope}, we employ the DIV2K dataset \cite{agustsson2017ntire} as training data. This dataset includes 800 images, each with a 2K resolution. For testing, we evaluate the methods on DIV2K validation and a wide range of standard benchmark datasets, including Urban100 \cite{huang2015single}, B100 \cite{martin2001database}, Set14 \cite{zeyde2010single} and Set5 \cite{bevilacqua2012low}.

{\bf Implementation details.} We  follow the experimental setup in previous ASISR methods \cite{chen2021learning, lee2022local, cao2023ciaosr, song2023ope} in this section. For training, we uniformly sample a scale factor $s$ from a continuous range of $[2, 4]$, and set the input LR image with size $48 \times 48$. 
The HR image patches are then randomly cropped from the data samples with size $48s \times 48s$, the paired LR image patches  are obtained by adopting the Bicubic downsampling on the HR patches for $s$ scale. Besides, we sample $48^2$ pixels and their corresponding coordinates from GT images to supervise the training process. 
For the training strategy, following the early method, we use Adam \cite{kingma2014adam} as the optimizer, and use L1 loss to train all models for 1000 epochs with a batch size of 16. We set the learning rate as 1e-4 at the beginning and decay to half of every 200 epochs. The experimental settings are the same as \cite{chen2021learning, lee2022local, cao2023ciaosr, song2023ope}. \red{Besides, when ameliorating CNN-based ASISR methods to their Rot-E versions, we follow the previous methods \cite{shen2020pdo,xie2022fourier} to set the convolution kernel size as $5\times 5$, which finely avoids the potential performance degradation caused by excessive reduction of parameters.}

{\bf Evaluation metrics.} We employ the Peak Signal-to-Noise Ratio (PSNR) to evaluate the quality of the synthesized SR images. Following \cite{chen2021learning, lee2022local}, the PSNR value is computed across all RGB channels for the DIV2K validation set. Additionally, for other benchmark test sets, the PSNR is calculated on the Y channel (luminance) of the transformed YCbCr space. 

{\bf In-scale SR results.} Table \ref{tab:in-scale} shows the in-scale ($\times 2, \times 3, \times 4$) super-resolution results on all the combinations of the 3 utilized  encoders and 3 utilized INRs. It can be easily observed that the proposed method consistently improves performance and reduces the number of parameters in all the backbone networks used. In particular, for the best backbone networks here, i.e., SwinIR-LTE, the proposed enhancement can further achieve a 0.2dB PSNR gain on Urban100 ($\times 2$). Additionally, as shown in Fig. \ref{fig:exp_in_scale}, we can also observe that the proposed method can better restore the texture details of the image.

\begin{figure}[ht]
	\centering
	\includegraphics[width=\linewidth]{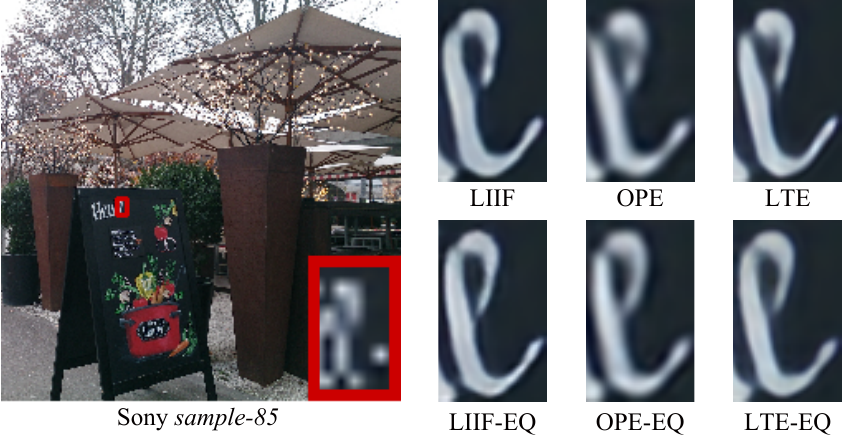} \vspace{-6mm}
	\caption{Visual comparison of different methods on a sample from  DPED\cite{ignatov2017dslr} datasets ($\times 8$ SR). All the models are trained on DIV2k dataset, and the encoder is EDSR.} \vspace{-2mm}
	\label{fig:exp_realsrset}
\end{figure}

\begin{table}[t]
	\centering\vspace{-0mm} 
	\caption{The generalization results on hyperspectral images. Average PSNR (dB) of the results obtained by all comparison methods on  CAVE \cite{yasuma2010generalized}.  }
	\centering \setlength{\tabcolsep}{5pt}\vspace{-0mm}
	\begin{tabular}{c c c c c c c c}
		\toprule
		 {Method} & {Mode}  & x2 & x3 & x4 & x6 & x8 & x12 \\
		\midrule
		 \multirow{2}{*}{LIIF \cite{chen2021learning}} 
		&Original& 44.75 & 41.26 & 39.08 & 36.25 & 34.50 & 31.93 \\
		&Proposed& \textbf{44.95} & \textbf{41.31} & \textbf{39.18} & \textbf{36.38} & \textbf{34.62} & \textbf{32.00} \\
		\cmidrule(r){1-2}\cmidrule(r){3-5}\cmidrule(r){6-8}
		 \multirow{2}{*}{OPE \cite{song2023ope}} 
		&Original& 44.96 & 41.25 & 39.09 & 36.22 & 34.33 & 31.71 \\
		&Proposed& \textbf{45.08} & \textbf{41.38} & \textbf{39.17} & \textbf{36.27} & \textbf{34.40} & \textbf{31.75} \\
		\cmidrule(r){1-2}\cmidrule(r){3-5}\cmidrule(r){6-8}
		 \multirow{2}{*}{LTE \cite{lee2022local}} 
		&Original& 45.05 & \textbf{41.45} & 39.21 & 36.30 & 34.47 & 31.80 \\
		&Proposed& \textbf{45.06} & \textbf{41.45} & \textbf{39.29} & \textbf{36.42} & \textbf{34.61} & \textbf{31.84} \\
		\bottomrule
	\end{tabular}
	\label{tab:cave}\vspace{-2mm}
\end{table}

{\bf Out-scale SR results.} Table \ref{tab:out-scale} shows the out-scale ($\times 6, \times 8, \times 12$) super-resolution results on all the combinations of the  3 utilized encoders and 3 utilized INR. It can be seen that the proposed method consistently  improves the performance in all the backbone networks utilized in the $\times 6$ and $\times 8$ cases. In the $\times 12$ case, the proposed framework achieves performance comparable to the original with fewer parameters. It worth nothing that in  $\times 12$ 
case, the super-resolved results are very blurry for the comparison methods,  making the PSNR results of the non-Rot-E and Rot-E methods similar to each other. Furthermore, as demonstrated in Fig. \ref{fig:exp_out_scale}, the proposed method exhibits superior capability in restoring fine texture details.

\red{It should be noted that the rotation equivariance amelioration  has very little impact on the inference speed. However, since we use $5\times 5$ convolution kernel, the inference time of CNN-based methods tend to be increased to a certain extent after our amelioration. Taking methods with LIIF-based INR as examples, the inference time (averaged on the texting set of Urban dataset, tested on RTX 4090 GPU) of EDSR-based and RDN-based method will increase from 0.018s to 0.037s and from 0.283s to 0.473s, respectively. Meanwhile the inference time of SwinIR-based methods changes very little (4.88s for original method, and 4.64s for proposed methods).}

\subsection{Real world Arbitrary-scale Image SR}
We extend the proposed method to real-world applications. Specifically, we train all the competing models on DIV2K dataset \cite{agustsson2017ntire}, and evaluate them on the DEPD dataset \cite{ignatov2017dslr}. As illustrated in Fig. \ref{fig:exp_realsrset}, although this experiment lacks ground truth for quantitative evaluation, visual comparisons still demonstrate
that the proposed Rot-E enhancement exhibits superior potential in practical scenarios.

\subsection{Generalization to Hyperspectral Image SR}
\begin{figure}
	\centering
	\includegraphics[width=1\linewidth]{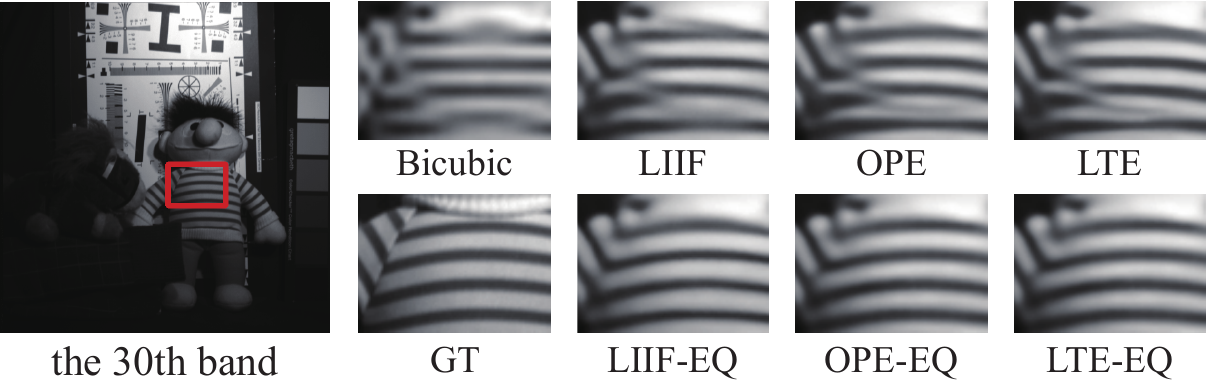} \vspace{-6mm}
	\caption{ Visual comparison of the super-resolution results ($\times 6$) from various methods on \textit{the 30-th band of chart and stuffed toy sample} from  CAVE  datasets. All models were trained on the DIV2K dataset using EDSR as encoder.}
    \vspace{-3mm}
	\label{fig:cave_ms}
\end{figure}

To further validate the generalization ability of the proposed method, we applied the completing models trained on the DIV2K natural image dataset \cite{agustsson2017ntire} to the super-resolution tasks on the CAVE hyperspectral image (HSI) dataset \cite{yasuma2010generalized}. 
Specifically, we test the performance on all the three aforementioned ASISR methods (i.e.,  LIIF \cite{chen2021learning}, OPE \cite{song2023ope} and LTE \cite{lee2022local}), and consistently set their encoder as the aforementioned EDSR-baseline \cite{lim2017enhanced} encoder.  Each band of the HSI is treated as grayscale images for processing SR individually.
As shown in Table \ref{tab:cave}, the proposed  Rot-E  methods improves the reconstruction performance on hyperspectral images compared to the original methods.
Additionally, Fig. \ref{fig:cave_ms} presents visualization results, illustrating that the proposed method substantially enhances texture restoration quality on hyperspectral images.

\subsection{\red{Generalization to Thermal Image SR}}
\begin{figure}[t]
	\centering
	\includegraphics[width=1\linewidth]{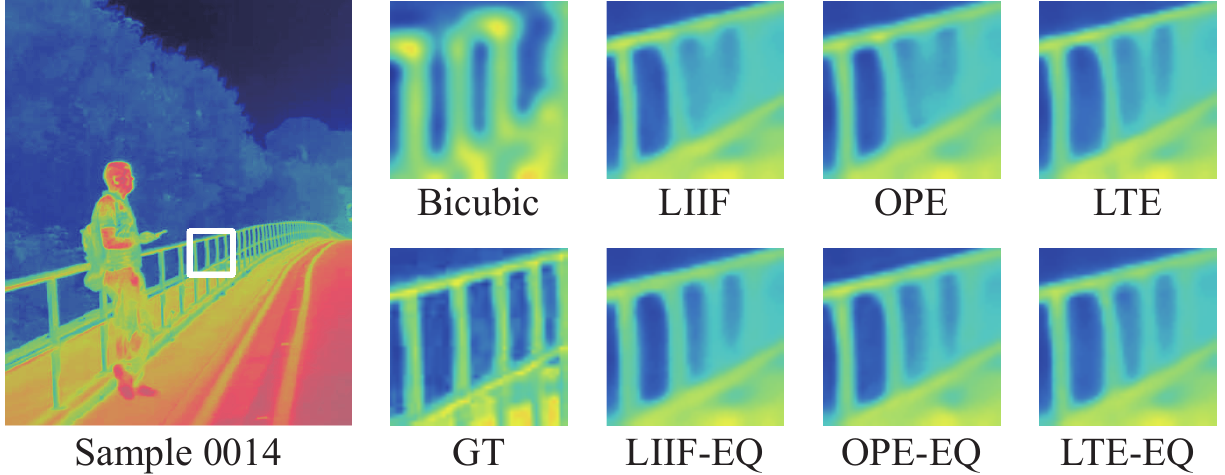} \vspace{-5mm}
	\caption{ Visual comparison of the super-resolution results ($\times 6$) from various methods on a sample from the thermal image dataset \cite{rivadeneira2019thermal}. All models were trained on the DIV2K dataset using EDSR as encoder.}
    \vspace{-3mm}
	\label{fig:infrared}
\end{figure}

\red{We also applied the trained models trained on DIV2k dataset to the thermal image dataset \cite{rivadeneira2019thermal}, for further texting the generalization ability. 
Similar as above experiment, we test the performance of all the three aforementioned ASISR methods, while consistently set their encoder as the aforementioned EDSR-baseline \cite{lim2017enhanced}.  
Table \ref{tab:infrared} demonstrates that the proposed Rot-E methods achieve superior reconstruction performance in thermal image super-resolution compared to the original approach. Fig. \ref{fig:infrared} provides visual evidence of the substantial quality improvement attained by our method.}

\begin{table}[t]
	\centering 
	\caption{\red{Generalization performance of the proposed method on thermal imagery: Comparative evaluation of average PSNR (dB)  obtained by all comparison methods on the testing set of thermal image dataset \cite{rivadeneira2019thermal}.}}
	\centering \setlength{\tabcolsep}{5pt}	\label{tab:infrared}\vspace{-2mm}
	\begin{tabular}{c c c c c c c c}
		\toprule
		 \red{Method} & \red{Mode}  & \red{x2} & \red{x3} & \red{x4} & \red{x6} & \red{x8} & \red{x12} \\
		\midrule
		 \multirow{2}{*}{\red{LIIF \cite{chen2021learning}}} 
		&\red{Original}& \red{43.02} & \red{38.17} & \red{35.28} & \red{32.05} & \red{30.33} & \red{28.13} \\
		&\red{Proposed}& \red{\textbf{43.10}} & \red{\textbf{38.30}} & \red{\textbf{35.38}} & \red{\textbf{32.08}} & \red{\textbf{30.37}} & \red{\textbf{28.17}} \\
		\cmidrule(r){1-2}\cmidrule(r){3-5}\cmidrule(r){6-8}
		 \multirow{2}{*}{\red{OPE \cite{song2023ope}}} 
		&\red{Original}& \red{\textbf{43.03}} & \red{38.20} & \red{35.36} & \red{32.12} & \red{30.37} & \red{\textbf{28.17}} \\
		&\red{Proposed}& \red{42.99} & \red{\textbf{38.24}} & \red{\textbf{35.40}} & \red{\textbf{32.15}} & \red{\textbf{30.39}} & \red{28.15} \\
		\cmidrule(r){1-2}\cmidrule(r){3-5}\cmidrule(r){6-8}
		 \multirow{2}{*}{\red{LTE \cite{lee2022local}}} 
		&\red{Original}& \red{42.96} & \red{38.15} & \red{35.28} & \red{32.03} & \red{30.31} & \red{28.13} \\
		&\red{Proposed}& \red{\textbf{43.07}} & \red{\textbf{38.27}} & \red{\textbf{35.38}} & \red{\textbf{32.12}} & \red{\textbf{30.39}} & \red{\textbf{28.15}} \\
		\bottomrule
	\end{tabular}
	\label{tab:infrared}\vspace{-0mm}
\end{table}

\begin{table*}[t]
  \caption{Average equivariant error on p4 rotation group under different settings. The results are averaged on 100 images in the DIV2K testing set.} \
  \label{tab:eqerror_ablation}
  \centering \setlength{\tabcolsep}{3.2pt}
  \begin{tabular}{cccccccccc}
    \toprule
     \multirow{3}{*}{ $\begin{array}{c}
        \mbox{Rot-E} \\
        \mbox{Encoder} 
      \end{array}$ } & \multirow{3}{*}{$\begin{array}{c}
        \mbox{Rot-E} \\
        \mbox{INR} 
      \end{array}$} &   \multicolumn{2}{c}{ $\times$2}  &  \multicolumn{2}{c}{ $\times$4} &  \multicolumn{2}{c}{ $\times$8} &  \multicolumn{2}{c}{ $\times$12}\\
    \cmidrule(r){3-4} \cmidrule(r){5-6} \cmidrule(r){7-8} \cmidrule(r){9-10}
    && NMSE $\downarrow$& NMAE$\downarrow$ & {NMSE $\downarrow$} & { NMAE $\downarrow$} & {NMSE $\downarrow$} & { NMAE $\downarrow$} & {NMSE $\downarrow$} & { NMAE $\downarrow$} \\
 \midrule  
   \ding{55} & \ding{55} & 0.019 $\!\pm\!$ 0.009 & 0.011 $\!\pm\!$ 0.006 & 0.030 $\!\pm\!$ 0.012 & 0.018 $\!\pm\!$ 0.008 & 0.039 $\!\pm\!$ 0.014 & 0.024 $\!\pm\!$ 0.010 & 0.041 $\!\pm\!$ 0.015 & 0.027 $\!\pm\!$ 0.011 \\ 
   \ding{51} & \ding{55} & 0.007 $\!\pm\!$ 0.003 & 0.005 $\!\pm\!$ 0.002 & 0.012 $\!\pm\!$ 0.005 & 0.007 $\!\pm\!$ 0.003 & 0.017 $\!\pm\!$ 0.007 & 0.011 $\!\pm\!$ 0.005 & 0.019 $\!\pm\!$ 0.008 & 0.012 $\!\pm\!$ 0.006 \\ 
   \ding{51} & \ding{51} & \textbf{5e-04 $\!\pm\!$ 5e-04} & \textbf{5e-05 $\!\pm\!$ 6e-05 }& \textbf{5e-04 $\!\pm\!$ 7e-04} & \textbf{6e-05 $\!\pm\!$ 8e-05 } & \textbf{7e-04 $\!\pm\!$ 9e-04} & \textbf{1e-04 $\!\pm\!$ 2e-04} & \textbf{8e-04 $\!\pm\!$ 1e-03} & \textbf{1e-04 $\!\pm\!$ 2e-04} \\ 
 \bottomrule  
\end{tabular}\vspace{-0mm}
\end{table*}

\begin{table*}[ht]
	\centering 
	\caption{Average PSNR (dB) of the super-resolution results obtained under different settings, on different benchmark datasets.}
	\centering \setlength{\tabcolsep}{7.9pt}
	\begin{tabular}{c c c c c c c c c c c c c c}
		\toprule
		 \multirow{3}{*}{ $\begin{array}{c}
                        \mbox{Rot-E} \\
                        \mbox{Encoder} 
                      \end{array}$ } & \multirow{3}{*}{$\begin{array}{c}
                        \mbox{Rot-E} \\
                        \mbox{INR} 
                      \end{array}$} & \multicolumn{3}{c}{Urban100 \cite{huang2015single}} & \multicolumn{3}{c}{BSD100 \cite{martin2001database}} & \multicolumn{3}{c}{Set14 \cite{zeyde2010single}} & \multicolumn{3}{c}{Set5 \cite{bevilacqua2012low}}\\
		\cmidrule(r){3-5}\cmidrule(r){6-8}\cmidrule(r){9-11}\cmidrule(r){12-14}
		& & x2 & x3 & x4 & x2 & x3 & x4 & x2 & x3 & x4 & x2 & x3 & x4 \\
		\midrule
		 \ding{55} & \ding{55} & 32.18 & 28.24 & 26.18 & 32.18 & 29.12 & 27.61 & 33.63 & 30.34 & 28.64 & 38.00 & 34.41 & 32.22 \\
          \ding{51} & \ding{55} & 32.34 & 28.34 & 26.26 & \textbf{32.22} & 29.14 & 27.62 & 33.65 & 30.35 & 28.66 & \textbf{38.01} & 34.42 & 32.20 \\
          \ding{51} & \ding{51} & \textbf{32.39} & \textbf{28.36} & \textbf{26.28} & {32.21} & \textbf{29.14} & \textbf{27.62} & \textbf{33.68} & \textbf{30.38} & \textbf{28.67} & \textbf{38.01} & \textbf{34.44} & \textbf{32.29} \\                           
		\bottomrule
	\end{tabular}
	\label{tab:eq_ablation_in_scale}
	\vspace{-0mm}
\end{table*}

\begin{table*}[ht]
	\centering 
	\caption{Ablation of filter parameterizations for Rot-E Convolutions. Average PSNR (dB) of the super-resolution results obtained by all comparison methods on different benchmark datasets.}
	\centering \setlength{\tabcolsep}{7.55pt}
	\begin{tabular}{c c c c c c c c c c c c c c}
		\toprule
		 \multirow{2}{*}{Method} & \multirow{2}{*}{Convlution} & \multicolumn{3}{c}{Urban100 \cite{huang2015single}} & \multicolumn{3}{c}{BSD100 \cite{martin2001database}} & \multicolumn{3}{c}{Set14 \cite{zeyde2010single}} & \multicolumn{3}{c}{Set5 \cite{bevilacqua2012low}}\\
		\cmidrule(r){3-5}\cmidrule(r){6-8}\cmidrule(r){9-11}\cmidrule(r){12-14}
		& & x2 & x3 & x4 & x2 & x3 & x4 & x2 & x3 & x4 & x2 & x3 & x4 \\
		\midrule

		 \multirow{3}{*}{LIIF \cite{chen2021learning}} 
		&CNN \cite{chen2021learning} & 32.18 & 28.24 & 26.18 & 32.18 & 29.12 & 27.61 & 33.63 & 30.34 & 28.64 & 38.00 & 34.41 & 32.22 \\
		&F-Conv\cite{xie2022fourier}& 32.28 & 28.26 & 26.18 & 32.19 & 29.11 & 27.59 & 33.64 & 30.33 & 28.61 & 37.97 & 34.38 & 32.21 \\
		&B-Conv (ours)& \textbf{32.39} & \textbf{28.36} & \textbf{26.28} & \textbf{32.21} & \textbf{29.14} & \textbf{27.62} & \textbf{33.68} & \textbf{30.28} & \textbf{28.67} & \textbf{38.01} & \textbf{34.44} & \textbf{32.29} \\
		\cmidrule(r){1-2}\cmidrule(r){3-5}\cmidrule(r){6-8}\cmidrule(r){9-11}\cmidrule(r){12-14}
		 \multirow{3}{*}{OPE \cite{song2023ope}} 
		&CNN \cite{song2023ope} & 31.93 & 28.06 & 25.93 & 32.09 & 29.03 & 27.52 & 33.54 & 30.24 & 28.53 & 37.90 & 34.17 & 31.98 \\
		&F-Conv\cite{xie2022fourier}& 31.94 & 28.00 & 25.85 & 32.11 & 29.04 & 27.51 & 33.54 & 30.26 & 28.50 & 37.88 & 34.24 & 31.99 \\
		&B-Conv (ours)& \textbf{32.00} & \textbf{28.15} & \textbf{25.98} & \textbf{32.14} & \textbf{29.06} & \textbf{27.55} & \textbf{33.55} & \textbf{30.27} & \textbf{28.56} & \textbf{37.95} & \textbf{34.28} & \textbf{32.06} \\
		\cmidrule(r){1-2}\cmidrule(r){3-5}\cmidrule(r){6-8}\cmidrule(r){9-11}\cmidrule(r){12-14}
		 \multirow{3}{*}{LTE \cite{lee2022local}} 
		&CNN \cite{lee2022local}& 32.28 & 28.32 & 26.23 & 32.21 & 29.14 & 27.62 & 33.68 & 30.38 & 28.66 & 38.03 & 34.50 & 32.27 \\
		&F-Conv \cite{xie2022fourier}& 32.27 & 28.25 & 26.18 & 32.21 & 29.13 & 27.61 & 33.65 & 30.36 & 28.65 & 38.02 & 34.42 & 32.21 \\
		&B-Conv (ours)& \textbf{32.43} & \textbf{28.37} & \textbf{26.31} & \textbf{32.23} & \textbf{29.17} & \textbf{27.64} & \textbf{33.74} & \textbf{30.42} & \textbf{28.69} & \textbf{38.07} & \textbf{34.51} & \textbf{32.28} \\
		
		\bottomrule
	\end{tabular}
	\label{tab:bconv_vs_fconv_in_scale}
	\vspace{-0mm}
\end{table*}


\subsection{Ablation Study}
{\bf Effectiveness of the equivariant modules.} In this part, we perform ablation studies to investigate the impact of the equivariance in each component of the proposed method. Specifically, we progressively replace the encoder and INR modules in the original ASISR method (encoder set as EDSR-baseline, INR module set as LIIF) with their equivariant versions. Table \ref{tab:eqerror_ablation} shows the rotation equivariant error in the p4 group, for the method under different settings. It can be easily observed that the method can only be rotation equivariant when it possesses both the Rot-E encoder and the proposed Rot-E INR.  Furthermore, as shown in Table \ref{tab:eq_ablation_in_scale}, the results demonstrate the effectiveness of the proposed modules in improving the SR performance.

\noindent
{\bf Different filter parameterizations for Rot-E convolutions.} Table \ref{tab:bconv_vs_fconv_in_scale} shows the comparison of the proposed filter parameterized method (B-Conv) and the SOTA parameterized equivariant convolution method, i.e., F-Conv\cite{xie2022fourier} (on all the tree aforementioned ASISR methods with their encoder consistently set as the aforementioned EDSR-baseline \cite{lim2017enhanced}). It can be clearly found that the proposed method has shown superior performance on all these benchmarks. More results can refer to the supplementary material.

\section{Conclusion}

In this work, we have introduced a novel formulation for constructing a rotation-equivariant framework for ASISR. We proposed a series of rotation-equivariant network modules for the INR in ASISR, which not only satifies the diverse design requirements for ASISR, but also enables the transformation of existing ASISR methods into their Rot-E counterparts, potentially enhancing performance while reducing the number of parameters. Our comprehensive theoretical analysis confirms that the rotation equivariance error of the proposed INR is exactly zero or approaches zero as the resolution of the input image increases. Additionally, we have developed a new filter parametrization method for constructing Rot-E convolutions, addressing inaccuracies present in current filter parametrization approaches. 

\red{Although the local features of most images have relatively strong rotational symmetry, there are still some special situations where the local features may not exhibit 360-degree rotational symmetry (e.g., rainy images). In such scenarios, introducing a strict rotational equivariance with  fixed-angle rotations equivariant inclines to hardly improve performance or even lead to a decline in performance.
Therefore, exploring equivariance with learnable transformations or partial angle rotations should be  an important research topic for the future. Besides, the Rot-E Transformer introduced in this study  was achieved by discarding positional encoding. Although we achieve $p4$ rotation equivariance, this improvement remains quite rudimentary.
Given the significant role of Transformers in recent image processing studies, future exploration of Rot-E Transformers could be of considerable importance.}

\bibliographystyle{unsrt}

\begin{IEEEbiography}[{\includegraphics[width=1in,height=1.25in,clip,keepaspectratio]{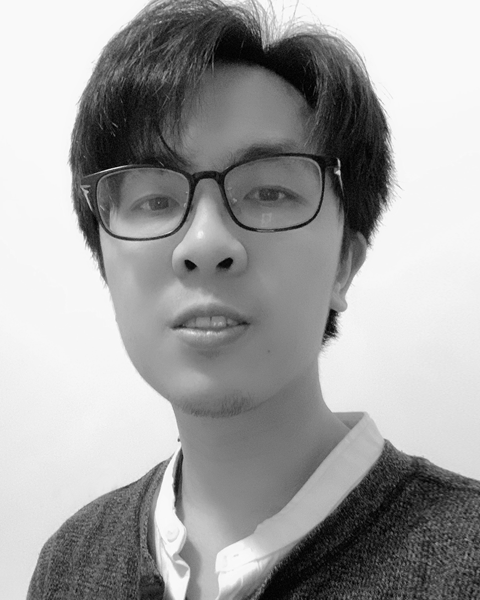}}]{Qi Xie} received the B.Sc. and Ph.D degree from Xi'an Jiaotong University, Xi'an, China, in 2013 and 2020 respectively.  Form 2018 to 2019, he was a Visiting Scholar in Princeton University, Princeton, NJ, USA. 
He is currently an associate professor in School of Mathematics and Statistics, Xi'an Jiaotong University.
His current research interests include model-based deep learning, filter parametrization-based deep learning, rotation equivariant deep learning.
\end{IEEEbiography}

\begin{IEEEbiography}[{\includegraphics[width=1in,height=1.25in,clip,keepaspectratio]{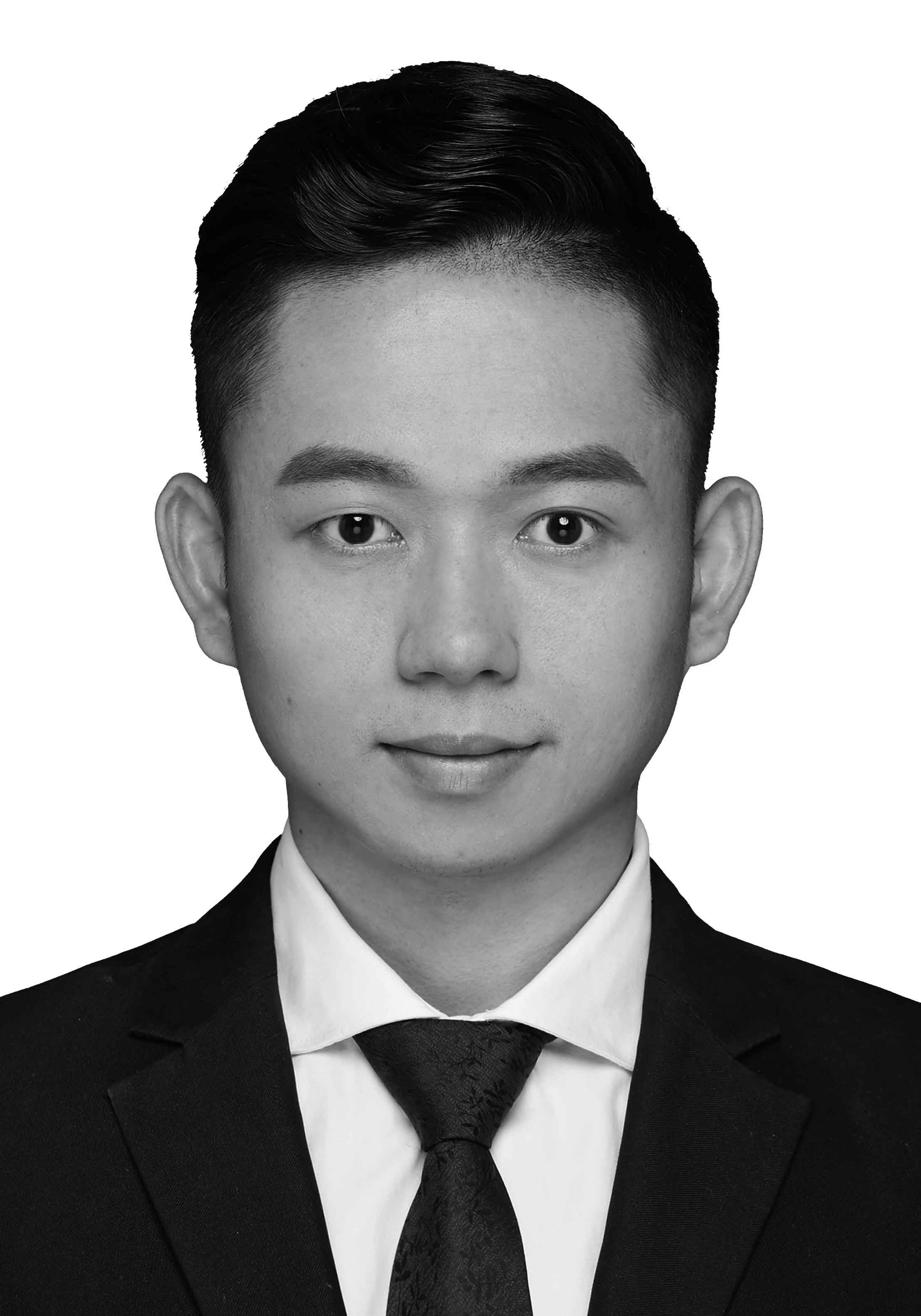}}]{Jiahong Fu} received the B.Sc. degree from North China Electric Power University, Beijing, China, in 2019. He is currently pursuing the Ph.D. degree in Xi'an Jiaotong University.
His current research interests include model-based deep learning and rotation equivariant deep learning.
\end{IEEEbiography}

\begin{IEEEbiography}[{\includegraphics[width=1in,height=1.25in,clip,keepaspectratio]{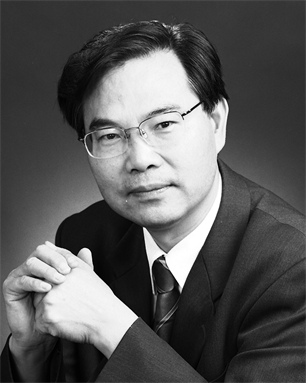}}]{Zongben Xu} received the Ph.D. degree in mathematics from Xi'an Jiaotong University, Xi'an, China, in 1987. He currently serves as the Academician of the Chinese Academy of Sciences, the Chief Scientist of the National Basic Research Program of China (973 Project), and the Director of the Institute for Information and System Sciences with Xi'an Jiaotong University. His current research interests include nonlinear functional analysis and intelligent information processing.
He was a recipient of the National Natural Science Award of China in 2007 and the winner of the CSIAM Su Buchin Applied Mathematics Prize in 2008.
\end{IEEEbiography}

\begin{IEEEbiography}[{\includegraphics[width=1in,height=1.25in,clip,keepaspectratio]{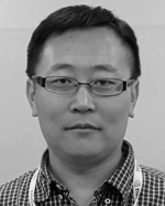}}]{Deyu Meng} received the B.Sc., M.Sc., and Ph.D. degrees from Xi'an Jiaotong University, Xi'an, China, in 2001, 2004, and 2008, respectively. He is currently a professor in School of Mathematics and Statistics, Xi'an Jiaotong University, and adjunct professor in Faculty of Information Technology, The Macau University of Science and Technology. From 2012 to 2014, he took his two-year sabbatical leave in Carnegie Mellon University. His current research interests include model-based deep learning, variational networks, and meta-learning.
\end{IEEEbiography}

\end{document}